\documentclass{article}

\usepackage{fontenc, inputenc, calc, indentfirst, fancyhdr, graphicx, epstopdf, lastpage, ifthen, lineno, float, amsmath, setspace, enumitem, mathpazo, booktabs, titlesec, etoolbox, tabto, xcolor, soul, multirow, microtype, tikz, totcount, changepage, attrib, upgreek, amsthm, hyphenat, hyperref, cleveref, footmisc, url, geometry, newfloat, caption}

\date{\vspace{-1cm}}

\usepackage{arydshln}

\title{Panoptic SwiftNet: Pyramidal Fusion for Real-time Panoptic Segmentation \thanks{The journal version of this article is available at \url{https://www.mdpi.com/2072-4292/15/8/1968}.}}

\author{
    Josip Šarić $^{1}$,  
    Marin Oršić $^{2}$ 
    and Siniša Šegvić $^{1}$}

\newcommand{\address}{{
  \footnotesize
  \begin{center}
  	\textit{University of Zagreb Faculty of Electrical Engineering and Computing, Unska 3, 10000 Zagreb\\
  	\{josip.saric, sinisa.segvic\}@fer.hr\\
  	Microblink Ltd., Strojarska cesta 20, 10000 Zagreb\\
  	marin.orsic@gmail.com}
  \end{center}
}}

\begin{document}
\maketitle
\address

\begin{abstract}
    Dense panoptic prediction 
is a key ingredient 
in many existing applications
such as autonomous driving,
automated warehouses or 
remote sensing.
Many of these applications
require fast inference 
over large input resolutions
on affordable or even embedded hardware.
We propose to achieve this goal
by trading off backbone capacity
for multi-scale feature extraction.
In comparison with contemporaneous 
approaches to panoptic segmentation,
the main novelties of our method
are efficient scale-equivariant 
feature extraction,
cross-scale upsampling 
through pyramidal fusion and 
boundary-aware learning 
of pixel-to-instance assignment.
The proposed method 
is very well suited 
for remote sensing imagery
due to the huge 
number of pixels 
in typical city-wide 
and region-wide datasets.
We present panoptic experiments
on Cityscapes, Vistas, COCO and
the BSB-Aerial dataset.
Our models outperform 
the state of the art 
on the BSB-Aerial dataset 
while being able to process
more than a hundred 1MPx images per second
on a RTX3090 GPU with FP16 precision
and TensorRT optimization.
\end{abstract}

\section{Introduction and related work}
Panoptic segmentation \cite{kirillov2019panoptic}
recently emerged as one of the most important
recognition tasks in computer vision.
It combines object detection \cite{ren2015faster}
and semantic segmentation \cite{long2015fully}
in a single task.
The goal is to assign a semantic class 
and instance index to each image pixel. 
Panoptic segmentation finds 
its applications in a
wide variety of fields
such as autonomous driving \cite{zendel2022unifying}, 
automated warehouses,
smart agriculture \cite{garnot2021panoptic},
and remote sensing \cite{osmar2022panoptic}.
Many of these applications
require efficient inference
in order to support timely decisions.
However,
most of the current state of the art
does not meet that requirement.
State-of-the-art panoptic approaches \cite{kirillov2019panopticfpn,xiong2019upsnet,Porzi_2019_CVPR,Hou_2020_CVPR}
are usually based on high-capacity backbones 
in order to ensure a
large receptive field
which is required for 
accurate recognition 
in large-resolution images.
Some of them are based on 
instance segmentation approaches
and therefore require
complex post-processing
in order to fuse 
instance-level detections
with pixel-level classification
\cite{kirillov2019panopticfpn,xiong2019upsnet}.

Different than all previous 
approaches to panoptic segmentation,
we propose to leverage 
multi-scale convolutional representations 
\cite{farabet12pami,orvsic2021efficient}
in order to increase the receptive field,
and decrease pressure 
onto the backbone capacity 
through scale equivariance
\cite{krevso2016convolutional}.
In simple words,
coarse input resolution
may provide a more appropriate 
view at large objects.
Furthermore, a mix of features 
from different scales is
both semantically rich
and positionally accurate.
This allows our model to achieve
competitive recognition performance 
and real-time inference 
on megapixel resolutions.

Early work on joint semantic 
and instance segmentation
evaluated model performance separately,
on both of the two tasks 
\cite{yao2012describing,dvornik2017blitznet}.
The field gained much attention
with the emergence 
of a unified task
called panoptic segmentation 
and the corresponding metric --- panoptic quality 
\cite{kirillov2019panoptic}.
This offered an easy transition 
path due to straight-forward
upgrade of the annotations:
panoptic segmentation labels
often can be created by combining 
existing semantic 
and instance segmentation ground truth.
Thus,
many popular recognition datasets 
\cite{Cordts_2016_CVPR,lin2014microsoft,neuhold2017mapillary}
were able to release
panoptic supervision
in a short amount of time,
which facilitated further research.

Most of the recent panoptic segmentation methods
fall under one of the two categories: 
box-based (or top-down) and
box-free (also known as bottom-up) methods.
Box-based methods predict the 
bounding box of the instance and 
the corresponding segmentation mask \cite{kirillov2019panopticfpn,xiong2019upsnet,mohan2021efficientps}.
Usually, semantic segmentation of stuff classes
is predicted by a parallel branch
of the model. 
The two branches have to be fused
in order to 
produce panoptic predictions.
Panoptic FPN \cite{kirillov2019panopticfpn}
extends the popular Mask R-CNN \cite{he2017mask}
with a semantic segmentation branch 
dedicated to the stuff classes.
Their fusion module
favors instance predictions
in pixels where both branches
predict a valid class.
Concurrent work \cite{xiong2019upsnet}
followed a similar extension idea,
but proposed a different post-processing step.
UPSNet \cite{xiong2019upsnet} 
stacks all instance segmentation masks
and stuff segmentation maps
and applies the softmax
to determine the instance index 
and semantic class in each pixel.
EfficientPS \cite{mohan2021efficientps}
combines EfficentNet backbone,
with custom upsampling 
and panoptic fusion.
Although, it 
improves the
efficiency
with respect to \cite{kirillov2019panopticfpn,mohan2021efficientps},
its inference speed 
is still far from real-time.

Box-free methods do not 
detect instances 
through bounding 
box regression \cite{yang2019deeperlab,Cheng_2020_CVPR,Li_2021_CVPR,Wang_2021_CVPR}.
Most of these methods
group pixels into instances
during a post-processing stage.
DeeperLab \cite{yang2019deeperlab}
extends a typical 
semantic segmentation architecture \cite{deeplabv3plus2018}
with a class-agnostic 
instance segmentation branch.
This branch detects
five keypoints per instance
and groups pixels
according to
multiple range-dependent
maps of offset vectors.
Panoptic Deeplab \cite{Cheng_2020_CVPR}
outputs a single heatmap of object centers
and a dense map of offset vectors 
which associates each thing pixel
with the centroid of 
the corresponding instance.
Panoptic FCN \cite{Li_2021_CVPR}
also detects
object centers,
however, they aggregate instance pixels
through regressed per-instance kernels.
Recent approaches \cite{Wang_2021_CVPR,cheng2021per,li2022maskdino}
propose unified frameworks
for all segmentation tasks,
including panoptic segmentation.
These frameworks include
special transformer-based modules
that recover
mask-level embeddings
and their
discriminative predictions.
Pixel embeddings 
are assigned to masks
according to the similarity
with respect to 
mask embeddings.
This approach seems inappropriate
for real-time processing of large images
due to large computational complexity
of mask-level recognition.

Semantic segmentation of remote-sensing images
is a popular and active line of research 
\cite{xu2020rsensing,liu2018rsensing,zhang2020rsensing,li22icip}.
On the other hand, 
only a few papers consider 
panoptic segmentation 
of remote-sensing images 
\cite{osmar2022panoptic, osmar2022rethinking, hua2021cascaded}.
Nevertheless,
it seems that panoptic segmentation
of remote-sensing images
might support many applications:
e.g.\ automatic compliance assessments
of construction work
with respect to urban legislation.
Just recently, 
a novel BSB aerial dataset
has been proposed \cite{osmar2022panoptic}.
The dataset collects
more than 3000 aerial images
of urban area in Brasilia
and the accompanying 
panoptic ground truth.
Their experiments involve Panoptic FPN \cite{kirillov2019panopticfpn}
with ResNet-50 and ResNet-101 backbones \cite{He_2016_CVPR}.
In contrast, our method combines
multi-scale ResNet-18 \cite{He_2016_CVPR} features
with pyramidal fusion \cite{orvsic2021efficient},
and delivers better PQ performance 
and significantly faster inference.

Multi-scale convolutional 
representations 
were introduced for semantic segmentation
\cite{farabet12pami}.
Further work \cite{orvsic2021efficient} showed
that they can outperform
spatial pyramid pooling 
\cite{zhao2017pyramid}
in terms of effective
receptive field.
To the best of our knowledge, 
this is the first work
that explores
the suitability of 
multi-scale representations
for panoptic segmentation.
Similar to \cite{Cheng_2020_CVPR}
our method relies on 
center and offset prediction
for instance segmentation.
However, our decoder
is shared between
instance and 
semantic segmentation branch.
Furthermore,
it learns with respect to 
a boundary-aware objective and
aggregates multi-scale features
using pyramidal fusion
in order to facilitate recognition
of large objects.

The summarized contributions of this work are as follows:
\begin{itemize}
    \item To the best of our knowledge,
    this is the first study of 
    multi-scale convolutional 
    representations \cite{farabet12pami}
    for panoptic segmentation.
    We show that pyramidal fusion
    of multi-scale convolutional
    representations significantly
    improves the panoptic segmentation
    in high-resolution images
    comparing to the single-scale variant.
    
    \item 
    We point out 
    that panoptic performance
    can be significantly improved
    by training pixel-to-instance assignment
    through a boundary-aware learning objective.
    
    \item Our models outperform
    generalization performance 
    of the previous state-of-the-art
    in efficient panoptic segmentation \cite{Cheng_2020_CVPR}
    while offering 
    around 60\% faster inference.
    Our method achieves
    state-of-the-art 
    panoptic performance
    on the BSB Aerial dataset
    while being
    at least 50\% faster
    than previous work \cite{osmar2022panoptic}.
\end{itemize}

\section{Method}
Our panoptic segmentation method
builds on scale-equivariant features \cite{farabet12pami},
and cross-scale blending 
through pyramidal fusion \cite{orvsic2021efficient}.
Our models deliver three dense predictions \cite{Cheng_2020_CVPR}:
semantic segmentation, instance centers
and offsets for pixel-to-instance assignment.
We attach all three heads
to the same latent representation 
at 4$\times$ subsampled resolution.
The regressed centers and offsets 
give rise to 
class-agnostic instances.
Panoptic segmentation maps are recovered
by fusing predictions
from all three heads \cite{Cheng_2020_CVPR}.

\subsection{Upsampling multi-scale features through pyramidal fusion}
We start from a ResNet-18 backbone \cite{He_2016_CVPR}
and apply it to a
three-level image pyramid 
that involves 
the original resolution 
as well as
$2\times$ and $4\times$ 
downsampled resolutions.
The resulting multi-scale representation
only marginally increases 
the computational effort
with respect to the single-scale baseline.
On the other hand,
it significantly increases 
the model's ability to recognize 
large objects in high-resolution images.
The proposed upsampling path 
effectively increases the receptive field
through cross-scale feature blending
which is also known as pyramidal fusion \cite{orvsic2021efficient}.
Pyramidal fusion proved 
beneficial for the 
semantic segmentation performance 
and we hypothesize that this
might be the case for 
panoptic segmentation as well.
In fact, a panoptic model 
needs to have a look 
at the whole instance 
in order to  stand a chance 
to recover the centroid
in spite of occlusions
and articulated motion.
This is hard to achieve
when targeting
real-time inference 
on large images 
since computational complexity 
is linear in the number of pixels.
We conjecture that pyramidal fusion
is an efficient way
to increase the receptive field
and enrich the features 
with the wider context.

\begin{figure}[h]
    \centering
    \includegraphics[width=\textwidth]{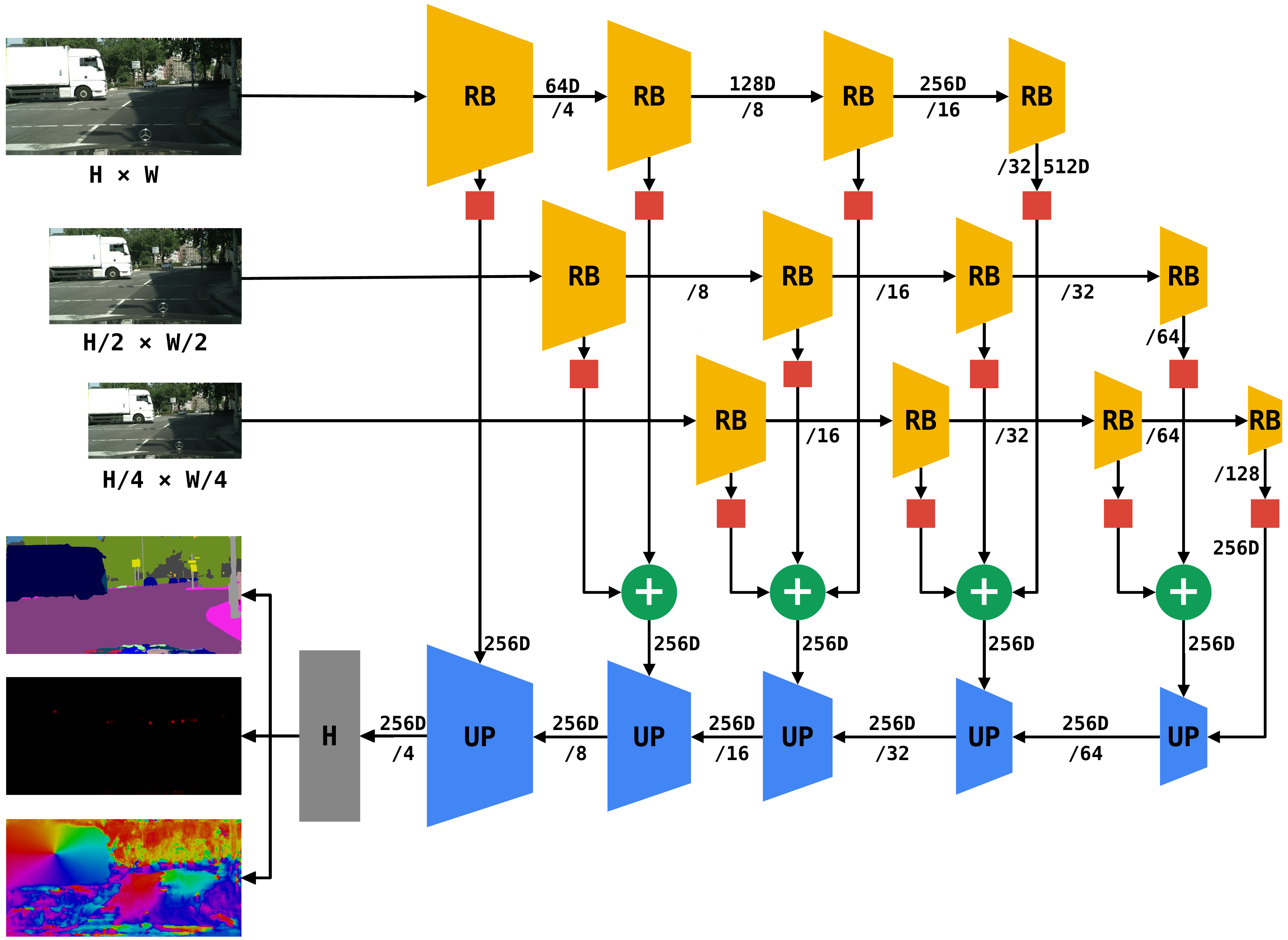}
    \caption{
             Panoptic SwiftNet with three-way 
             multi-scale feature extraction, 
             pyramidal fusion, 
             common upsampling path, 
             and three prediction heads.
             Yellow trapezoids denote residual blocks (RB).
             Red squares represent a $1\times1$
             convolutions that adjust 
             the number of feature maps
             so that the skip connection 
             can be added  to the upsampling stream.
             Blue trapezoids represent upsampling modules. 
             The gray rectangle represents
             the three prediction heads.
             Modules with the same color share parameters.
             Numbers above the lines (ND) 
             specify the number of channels.
             Numbers below the lines (/N)
             specify the downsampling factor
             w.r.t.\ to the original resolution
             of the input image.
             }
    \label{fig:psn_model}
\end{figure}

Figure~\ref{fig:psn_model}
illustrates the resulting architecture.
Yellow trapezoids represent
residual blocks of a typical backbone
which operate on 
$4\times$, $8\times$, $16\times$ and $32\times$ 
subsampled resolutions.
The trapezoid color 
indicates feature sharing: 
the same instance of the 
backbone is applied
to each of the input images.
Thus our architecture complements scale-equivariant
feature extraction with scale-aware upsampling.
Skip connections from the residual blocks
are projected with a single $1\times1$
convolution (red squares)
in order to match the
number of channels
in the upsampling path.
Features from different 
pyramid levels
are combined with
elementwise addition 
(green circles).
The upsampling path consists of 
five upsampling modules.
Each module fuses 
features coming from 
the backbone via the 
skip connections 
and features from 
the previous upsampling stage.
The fusion is performed through
elementwise addition and
a single $3\times3$ convolution.
Subsequently, the fused features 
are upsampled
by bilinear interpolation.
The last upsampling module outputs
a feature tensor which is
four times subsampled
w.r.t.\ to the input resolution.
This feature tensor
represents a shared input
to the three dense prediction heads
that recover the 
following three tensors:
semantic segmentation,
centers heatmap
and offset vectors.
Each head consists of
a single $1\times1$ convolution
and $4\times$ bilinear upsampling.
Such design of the prediction heads
is significantly faster 
than in Panoptic Deeplab \cite{Cheng_2020_CVPR},
because it avoids inefficient
depthwise-separable convolutions 
with large kernels on 
large resolution inputs.
Please note that each convolution in the above analysis
actually corresponds to a BN-ReLU-Conv unit \cite{he2016identity}.

The proposed upsampling path has to provide both 
instance-agnostic features for semantic segmentation
as well as instance-aware features for panoptic regression.
Consequently, we use 256 channels 
along the upsampling path,
which corresponds to twice larger capacity
than in efficient semantic models 
\cite{orvsic2021efficient}.

The proposed design differs 
from the previous work
which advocates separate
upsampling paths for semantic 
and instance-specific predictions  
\cite{Cheng_2020_CVPR}.
Our preliminary validation
showed that shared intermediate features
can improve panoptic performance
when there is enough upsampling capacity.

\subsection{Boundary-Aware Learning of Pixel-to-Instance Assignment}
\label{sec:baol}
Regression of offset centers requires
large prediction changes at instance boundaries.
A displacement of only one pixel
must make a distinction between two
completely different instance centers.
We conjecture that such abrupt changes 
require a lot of capacity. 
Furthermore, it is intuitively clear
that the difficulty of offset regression
is inversely proportional to the distance
from the  instance boundary. 

Hence, we propose to prioritize 
the pixels at instance boundaries
by learning offset regression  
through boundary-aware loss \cite{orvsic2021efficient,zhen2019learning}. 
We divide each instance into four regions 
according to the distance from the boundary. 
Each of the four regions is assigned 
a different weight factor.
The largest weight factor is assigned
to the regions which are closest to the border.
The weights diminish as we go
towards the interior of the instance.
Consequently, we formulate 
the boundary-aware offset loss
as follows:
\begin{align}
    \mathcal{L}_{BAOL} = \frac{1}{H \cdot W} \sum_{i,j}^{H,W}{w_{i,j} \cdot \left|O_{i,j} - O_{i,j}^\textrm{GT}\right|} \ ,\  w_{i,j} \in \{1, 2, 4, 8\} \label{eq:bal}
\end{align}
In the above equation
H and W represent 
height and width 
of the input image,
$O_{i,j}$ and $O_{i,j}^\textrm{GT}$
--- predicted and ground-truth
offset vectors at location $(i,j)$,
and $w_{i,j}$ ---
boundary-aware weight at the same location.

Note that this is quite different than 
in earlier works \cite{orvsic2021efficient,zhen2019learning} 
which are all based on focal loss
\cite{lin2017focal}.
To the best of our knowledge,
this is the first use 
of boundary modulation
in a learning objective
based on L1 regression. 

\subsection{Compound learning objective}
We train our models
with the compound loss
consisting of three components.
The semantic segmentation loss $\mathcal{L}_{\mathrm{SEM}}$
is expressed as 
the usual per-pixel 
cross entropy \cite{long2015fully}.
In Cityscapes experiments 
we additionally use
online hard-pixel mining 
and consider only the
top $20\%$ pixels 
with the largest loss \cite{deeplabv3plus2018}.
The center regression loss $\mathcal{L}_{\mathrm{CEN}}$
corresponds to L2 loss 
between the predicted centers heatmap
and the ground-truth heatmap \cite{Cheng_2020_CVPR}.
Such learning objective
is common in learning 
heatmap regression for 
keypoint detection 
\cite{xiao2018simple,newell2016stacked}.
The offset loss $\mathcal{L}_{\mathrm{BAOL}}$
corresponds to the modulated L1 loss
as described in (\ref{eq:bal}).
The three losses are modulated
with hyperparameters $\lambda_\mathrm{SEM}$, $\lambda_{\mathrm{CEN}}$, and $\lambda_{\mathrm{BAOL}}$:
\begin{align}
    \mathcal{L} = \lambda_{\mathrm{SEM}} \cdot \mathcal{L}_{\mathrm{SEM}} + \lambda_{\mathrm{CEN}} \cdot \mathcal{L}_{\mathrm{CEN}} + \lambda_{\mathrm{BAOL}} \cdot \mathcal{L}_{\mathrm{BAOL}}
    \label{eq:comp_loss}
\end{align}
We express these hyper-parameters 
relative to the contribution 
of the segmentation loss 
by setting $\lambda_\mathrm{SEM}=1$.  
We set $\lambda_{\mathrm{CEN}}=200$ in early experiments
so that the contribution of the center loss 
becomes approximately equal to
the contribution of the segmentation loss.
We validate $\lambda_{\mathrm{BAOL}}=0.0025$ 
in early experiments
on the Cityscapes dataset.

\begin{figure}[h]
    \centering
    \begin{tabular}{@{}c@{\,}c@{\,}c@{}}
        \multirow{2}{*}[2.2cm]{\includegraphics[height=0.4\textwidth,width=0.4\textwidth]{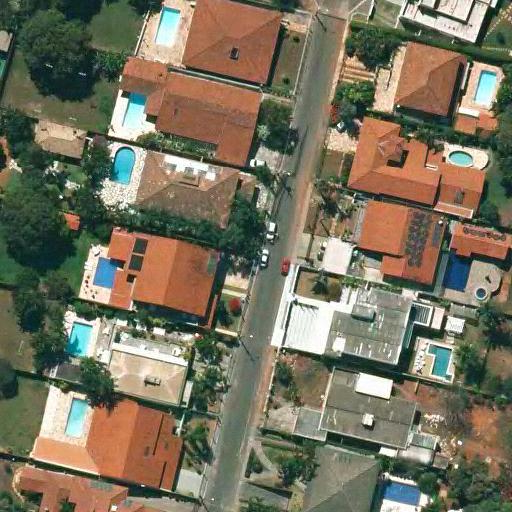}} & 
        \includegraphics[width=0.3\textwidth]{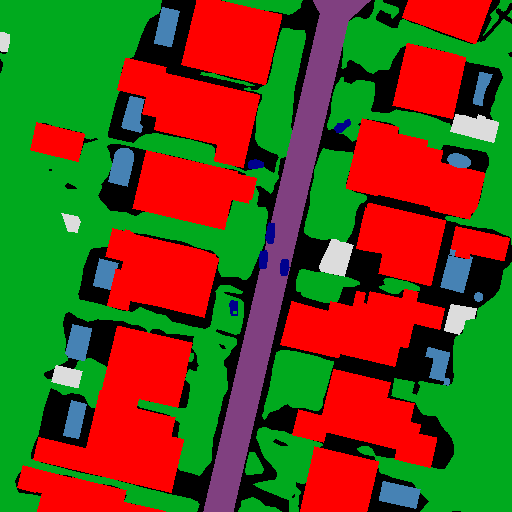} & 
        \includegraphics[width=0.3\textwidth]{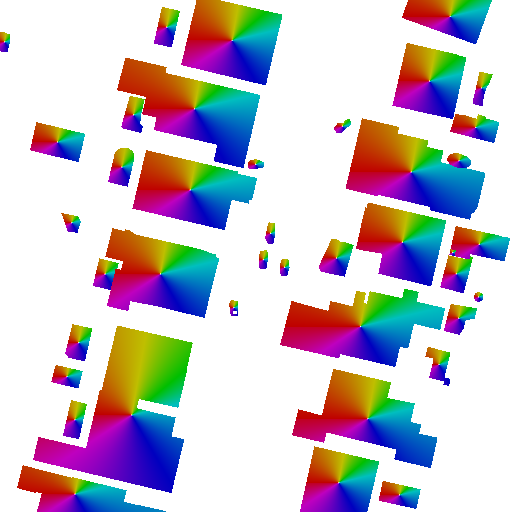} \\[-0.2em]
        & \includegraphics[width=0.3\textwidth]{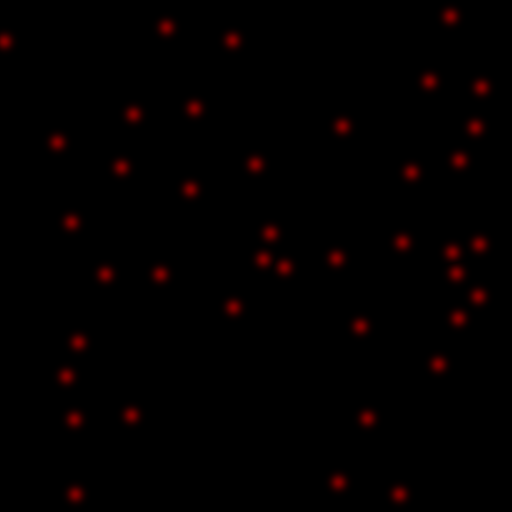} & 
        \includegraphics[width=0.3\textwidth]{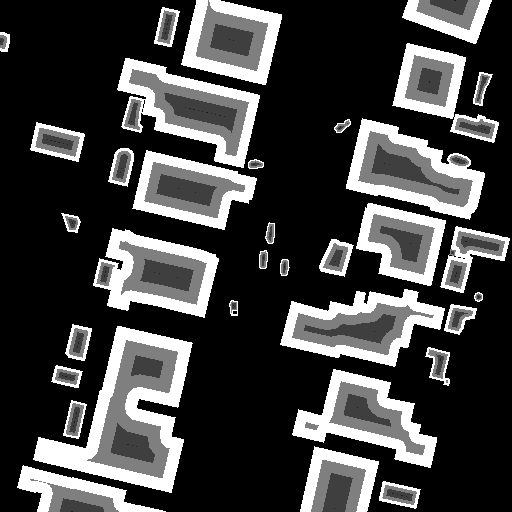} \\
        (a) & \multicolumn{2}{c}{(b)} \\
    \end{tabular}
    \caption{A Panoptic Swiftnet training example consists of an input image (a) and the corresponding ground-truth labels (b):
    semantic segmentation (top left), offset vectors (top right), heatmap for instance centers (bottom left), and weights for boundary-aware offset loss (bottom right). 
    Offset directions are color coded
    according to the convention from the previous work
    in optical flow \cite{baker2011database}.
    }
    \label{fig:gt_labels}
\end{figure}

Figure~\ref{fig:gt_labels} (b) shows
ground truth targets for a single 
training example on the BSB Aerial dataset.
The semantic segmentation labels 
associate each pixel 
with categorical semantic ground truth (top-left).
Ground-truth offsets point towards
the respective instance centers.
Notice abrupt changes at 
instance-to-instance boundaries (top-right).
The ground-truth instance-center heatmap is crafted 
by Gaussian convolution ($\sigma=5$) of a binary image
where ones correspond to instance centers (bottom-left).
We craft the offset-weight ground-truth
by thresholding the distance transform \cite{DBLP:journals/cvgip/Borgefors86}
with respect to instance boundaries.
Note that pixels at stuff classes 
do not contribute to the offset loss
since the corresponding weights 
are set to zero (bottom-right).

\subsection{Recovering panoptic predictions}
We recover panoptic predictions through
post-processing of the three model outputs
as in Panoptic Deeplab \cite{Cheng_2020_CVPR}.
We first recover instance centers
by non-maximal suppression
of the center heatmap.
Second, each thing pixel is assigned
the closest instance center
according to the corresponding
displacement from the offset map.
Third, each instance 
is assigned a semantic class
by taking arg-max over
the corresponding semantic histogram.
This voting process presents an opportunity
to improve the original semantic predictions,
as demonstrated in Figure 4 (col.\ 1).
Finally, the panoptic map is constructed
by associating instance indices from step 2
with aggregated semantic evidence from step 3.

We greatly improve the execution speed
of our Python implementation
by implementing custom CUDA modules
for steps 3 and 4.
Still, the resulting CUDA implementation 
requires approximately the same time 
as optimized TensorRT inference
in our experiments on Jetson AGX.
This suggests that further improvement 
of the inference speed may require
rethinking of the post-processing step.

\section{Experiments}
We consider multi-scale models
with pyramidal fusion 
based on ResNet-18 \cite{He_2016_CVPR}.
We evaluate the performance
of our panoptic models
with standard metrics 
\cite{kirillov2019panoptic}:
panoptic quality (PQ),
segmentation quality (SQ),
and recognition quality (RQ).
In some experiments,
we also evaluate our panoptic models
on semantic and instance segmentation tasks,
and show the corresponding performance in 
terms of mean intersection over union (mIoU) 
and average precision (AP).
For completeness,
we now briefly review
the used metrics.

\newcommand{\TP}{\mathit{TP}}
\newcommand{\FN}{\mathit{FN}}
\newcommand{\FP}{\mathit{FP}}
\newcommand{\IoU}{\text{IoU}}
\newcommand{\mIoU}{\text{mIoU}}

Mean intersection over union (mIoU)
measures the quality
of semantic segmentation.
Let $P_c$ and $G_c$
denote all pixels 
that are predicted and
labeled as class $c$.
Then,
equation (\ref{eq:miou})
defines 
$\IoU_c$ as ratio
between the
intersection 
and the union
of $P_c$ and $G_c$.
This ratio is
also known as Jaccard index.
Finally,
the mIoU metric is simply 
an average over 
all classes $\mathcal{C}$.

\begin{align}
    \label{eq:miou}
    \IoU_c = \frac{|P_c \cap G_c|}{|P_c \cup G_c|}
    \quad 
    \mIoU = \frac{\sum_{c \in \mathcal{C}}\IoU_c}{|\mathcal{C}|} 
\end{align}

Panoptic quality (PQ)
measures the similarity
between the predicted
and the ground truth
panoptic segments
\cite{kirillov2019panoptic}.
Each panoptic segment collects
all pixels labeled with
the same semantic class
and instance index.
As in intersection over union,
PQ is first computed
for each class separately,
and then averaged 
over all classes.
In order to compute 
the panoptic quality,
we first need to 
match each ground 
truth segment $g$
with the predicted segment $p$.
The segments are matched
if they overlap
with $\IoU > .5$.
Thus, 
matched predicted segments
are considered 
as true positives (TP),
unmatched predicted segments
as false positives (FP),
and unmatched ground truth segments
as false negatives (FN).
The equation (\ref{eq:pq}) shows
that PQ is proportional 
to the average $\IoU$
of true positives,
and inversely proportional
to the number of false positives
and false negatives.
\begin{align}
    \label{eq:pq}
     {\text{PQ} = \frac{\sum_{(p, g) \in \TP} \text{IoU}(p, g)}{|\TP| + \frac{1}{2}|\FP| + \frac{1}{2}|\FN|} \,.}
\end{align}

PQ can be further factorized
into segmentation quality (SQ)
and recognition quality (RQ) \cite{kirillov2019panoptic}.
This can be achieved
by multiplying the eq.\ (\ref{eq:pq})
with $\frac{|\TP|}{|\TP|}$
as shown in the equation (\ref{eq:sqrq}).
The equation clearly indicates that
the segmentation quality 
corresponds to the average $\IoU$
of the true positive segment pairs,
while the recognition quality
corresponds to the F1 score
of segment detection
for the particular class.
\begin{align}
    \label{eq:sqrq}
    \small{\text{PQ}} = \underbrace{\frac{\sum_{(p, g) \in \TP} \text{IoU}(p, g)}{\vphantom{\frac{1}{2}}|\TP|}}_{\text{segmentation quality (SQ)}} \times \underbrace{\frac{|\TP|}{|\TP| + \frac{1}{2} |\FP| + \frac{1}{2} |\FN|}}_{\text{recognition quality (RQ) }} \,.
\end{align}

Finally, we briefly recap
the average precision (AP)
for instance segmentation.
In recent years,
this name has become a synonym
for COCO mean average precision 
\cite{lin2014microsoft}.
This metric averages 
the traditional AP over 
all possible classes
and 10 different IoU thresholds 
(from 0.5 to 0.95).
In order to compute 
the traditional AP,
each instance prediction
needs to be associated
with a prediction score.
The AP measures the 
quality of ranking 
induced by the prediction score.
It is then computed
as the area under the
precision-recall curve
which is obtained by
considering 
all possible thresholds 
of the prediction score.
Each prediction
is considered a true positive
if it overlaps some
ground truth instance
more than the selected 
IoU threshold.

Our extensive 
experimental study
considers four 
different datasets.
We first present our results
on the BSB Aerial dataset 
\cite{osmar2022panoptic}
which collects
aerial images.
Then, we evaluate 
our models on 
two road-driving datasets 
with high-resolution images: 
Cityscapes \cite{Cordts_2016_CVPR} 
and Mapillary Vistas \cite{neuhold2017mapillary}.
Finally, we present an evaluation on the COCO dataset \cite{lin2014microsoft} 
which gathers a very large number of 
medium-resolution images 
from personal collections.
Moreover,
we provide a validation study
of our panoptic model
on Cityscapes.
We chose Cityscapes 
for this study
because of its 
appropriate size
and high-resolution 
images which provide 
a convenient challenge
for our pyramidal fusion design.
In the end, 
we measure the inference speed 
of our models on different GPUs
with and without TensorRT optimization.

\subsection{BSB Aerial dataset}
The BSB Aerial dataset
contains 3000 training,
200 validation
and 200 test images
of urban areas in
Brasilia, Brazil.
All images have the same
resolution of $512 \times 512$ pixels
and are densely labeled with  
3 stuff (street, permeable area, and lake) 
and 11 thing classes (swimming pool, harbor, vehicle, boat, sports court, soccer field, commercial building, commercial building block, residential building, house, and small construction).
The highest number
of pixels is annotated
as 'permeable area' 
because this class considers
different types of
natural soil and vegetation.

We train our models for 120000 iterations
on batches consisting of 24 
random crops with resolution 
$512 \times 512$ pixels.
We use ADAM \cite{kingma15adam} optimizer
with a base learning rate 
of $3 \cdot 10^{-4}$
which we polynomially decay
to $10^{-7}$.
We augment input images
with random scale jitter
and horizontal flipping.
We also validate
image rotation 
for data augmentation 
and present the outcomes 
in Table \ref{tab:bsb_rot}.

Table~\ref{tab:bsb_main} compares our 
experimental performance
on BSB Aerial dataset 
with the related work.
Our model based on ResNet-18
outperforms Panoptic FPN
with stronger backbones
by a large margin.
We note improvements over Panoptic FPN 
with ResNet-50 and ResNet-101
of 5.7 and 4.2 PQ points on the validation data,
and 5.9 and 3.1 PQ points on the test data.
Besides being more accurate,
our model is also 
significantly faster.
In fact, our model is 
1.5 times faster than
untrained Panoptic-FPN-ResNet-50.

\begin{table}[h]
    \centering
        \caption{Performance evaluation of panoptic segmentation methods $512\times512$ images from the BSB Aerial dataset. FPS denotes frames per second from PyTorch and Python.
        $\dagger$ - denotes our estimates that measure the inference speed of uninitialized models which on average detect less than 1 instance across the validation dataset.}
    \begin{tabular}{lccccccc}
         Model & PQ & SQ & RQ & FPS & GPU \\
         \toprule
         \multicolumn{5}{c}{VALIDATION SET} \\ 
         \toprule
         PanopticFPN-R50 \cite{osmar2022panoptic} & 63.8 & 84.9 & 74.6 & 41.3 $\dagger$ & RTX3090\\
         PanopticFPN-R101 \cite{osmar2022panoptic} & 65.3 & 85.1 & 76.2 & 31.8 $\dagger$ & RTX3090\\
         PanopticSwiftNet-RN18 (ours) & \textbf{69.5} & \textbf{89.3} & \textbf{77.5} & \textbf{64.9} & RTX3090 \\
         \toprule
         \multicolumn{5}{c}{TEST SET} \\ 
         \toprule
         SemSeg-Hybrid \cite{osmar2022rethinking} & 47.4 & 81.3 & 58.3 & - & - \\
         PanopticFPN-R50 \cite{osmar2022panoptic} & 62.2 & 85.3 & 72.2 & 41.3 $\dagger$ & RTX3090\\
         PanopticFPN-R101 \cite{osmar2022panoptic} & 65.0 & 85.4 & 75.5 & 31.8 $\dagger$ & RTX3090\\
         PanopticSwiftNet-RN18 (ours) & \textbf{68.1} & \textbf{88.2} & \textbf{76.0} & \textbf{64.9} & RTX3090 \\
    \end{tabular}
    \label{tab:bsb_main}
\end{table}
Note that 
we estimate the
inference speed
of Panoptic FPN
with a randomly 
initialized model
which detects less than 
one instance per image.

Image rotation is rarely used
for data augmentation on typical
road-driving datasets.
It makes little sense
to encourage the models
robustness to rotation
if the camera pose is nearly fixed 
and consistent across all images
from both train and validation subsets.
However, it seems that 
rotation in aerial images
could simulate other 
possible poses
of the acquisition platform,
and thus favor better 
generalization on test data.
Table 2 validates this hypothesis 
on the BSB aerial dataset.
We train our model in 
three different setups.
The first setup does not
use rotation.
The second rotates each 
training example
for a randomly 
chosen angle
from the set
$\{0^\circ, 90^\circ, 180^\circ, 270^\circ\}$.
The third setup rotates
each training example
for an angle
randomly sampled
from a range of 
$0^\circ - 360^\circ$.
Somewhat surprisingly, 
the results 
show that 
rotated inputs 
decrease the generalization
ability on BSB val.
The effect is even stronger 
when we sample 
arbitrary rotations.
Visual inspection suggests that this happens 
due to the constrained acquisition process.
In particular, we have noticed
that in all inspected images
the shadow is  
pointing at roughly 
the same direction.
We hypothesize that 
the model learns 
to use this fact
as some sort of 
orientation landmark
for offset prediction.
This is similar to
road driving scenarios
where the models usually 
learn the bias
that the sky covers
the top part of the image
while the road 
covers the bottom.
Clearly,
rotation of the training examples
prevents the model
in learning these biases,
which can hurt 
the performance
on the validation set.
The experiments support
this hypothesis
because the deterioration
is much smaller in
semantic performance than
in panoptic performance.
We remind the reader that,
unlike panoptic segmentation,
semantic segmentation 
does not require
offset predictions.

\vspace{0.3cm}
\begin{table}[h]
    \centering
        \caption{Validation of rotation for data augmentation on the validation set of
        the BSB Aerial dataset.}
    \begin{tabular}{lcccc}
         Rotation policy & PQ & SQ & RQ & mIoU \\
         \toprule
          None 
          & 69.5 & 89.3 & 77.5 & 75.1 \\
          Angle $\in \{0^\circ, 90^\circ, 180^\circ, 270^\circ\}$ 
          & 68.8 & 87.8 & 77.6 & 74.8 \\
          Angle $\in [0^\circ, 360^\circ]$
          & 56.0 & 86.1 & 63.9 & 74.5 \\
    \end{tabular}
    \label{tab:bsb_rot}
\end{table}
\vspace{0.3cm}

Figure~\ref{fig:bsb_preds}
shows the predictions
of our model
on four scenes
from BSB Aerial val.
Rows show: 
input image, 
semantic segmentation, 
centers heatmap, 
offset directions, 
and panoptic segmentation.
Panoptic maps designate instances 
with different shades of the color
of the corresponding class.

The presented scenes illustrate 
the diversity of the BSB dataset.
We can notice 
a large green area 
near the lake (col.\ 1),
but also 
urban area
with streets and houses
(col.\ 4).
There is 
a great variability
in object size.
For example,
some cars in column 2
are only 10 pixels wide,
while the building
in column 3 is nearly 
400 pixels wide.
Our model deals
with this variety 
quite well.
We can notice
that we correctly 
detect and segment
most of the cars,
and also larger objects
such as buildings.
However,
in the first column
we can notice that
most of the soccer field
in top-right 
part of the image
is mistakenly segmented
as permeable area.
Interestingly,
center detection
and offset prediction
seem quite accurate,
but semantic segmentation
failed to discriminate
between the two classes.
This is a fairly 
common and reasonable mistake
because the two classes
are visually similar.
In the last column,
we notice that the
road segments 
are not connected.
However, this is caused by
the labeling policy
which considers trees 
above the road
as permeable area
or unlabeled.

\newcommand{\mywbsb}{0.24\textwidth}
\begin{figure}[h!]
    \centering
    \begin{tabular}{@{}c@{\,}c@{\,}c@{\,}c@{}}
            \includegraphics[width=\mywbsb]{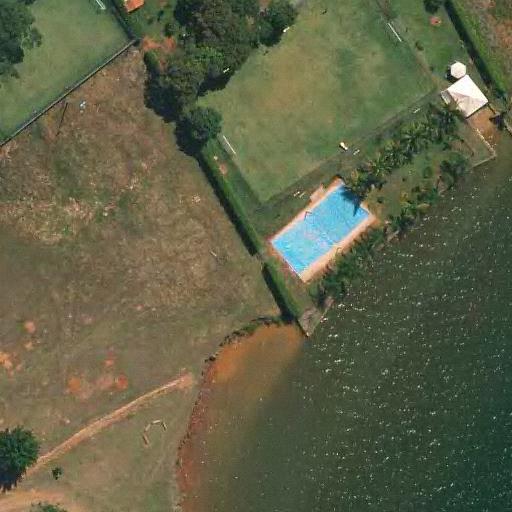} & 
            \includegraphics[width=\mywbsb]{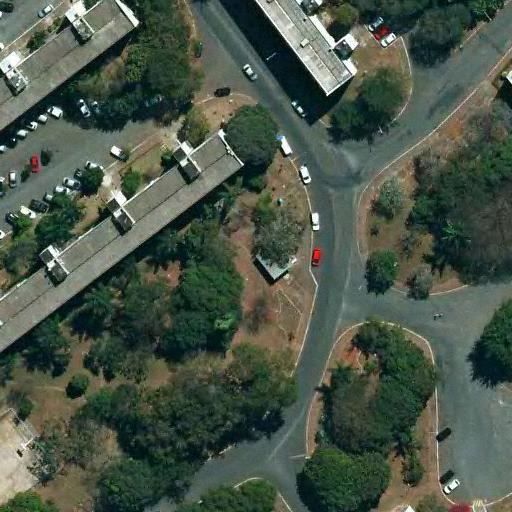} & 
            \includegraphics[width=\mywbsb]{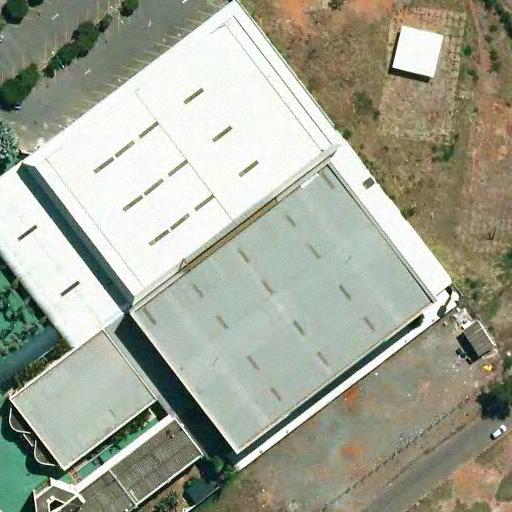} & 
            \includegraphics[width=\mywbsb]{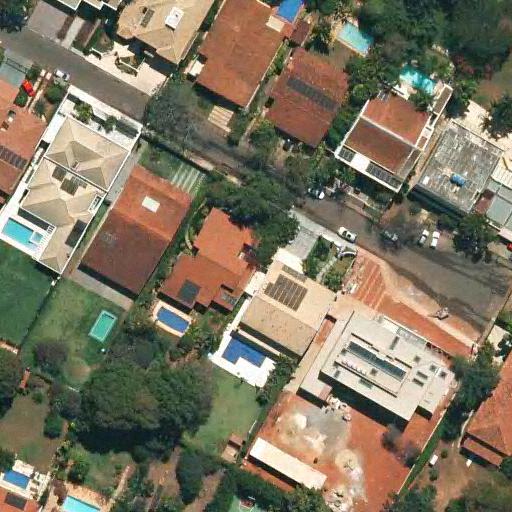}
            \\[-0.2em]

            \includegraphics[width=\mywbsb]{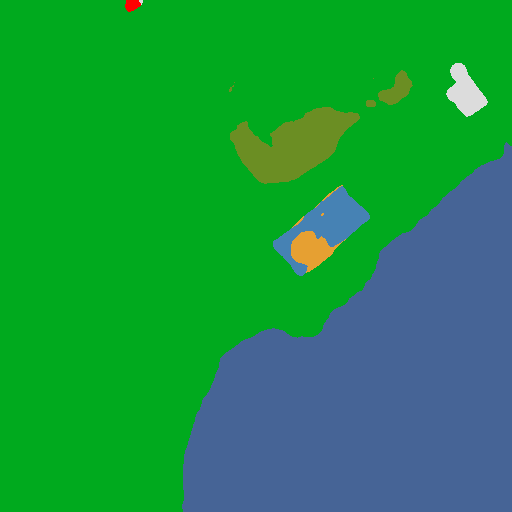} & 
            \includegraphics[width=\mywbsb]{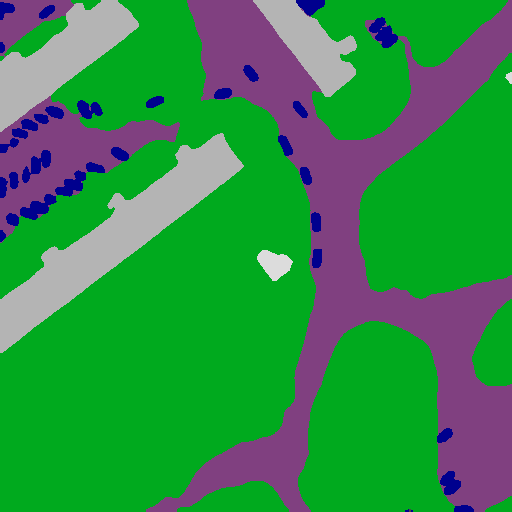} & 
            \includegraphics[width=\mywbsb]{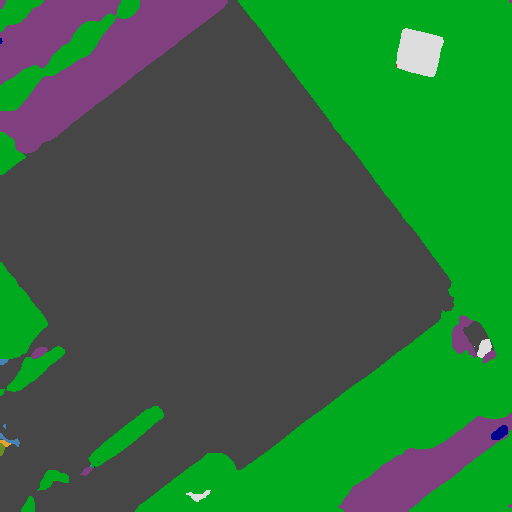} & 
            \includegraphics[width=\mywbsb]{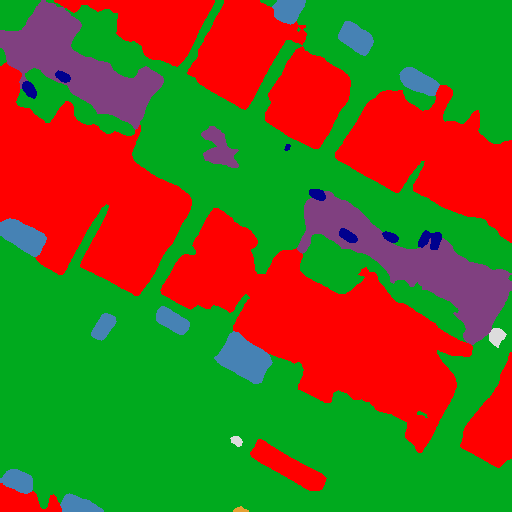}
            \\[-0.2em]

            \includegraphics[width=\mywbsb]{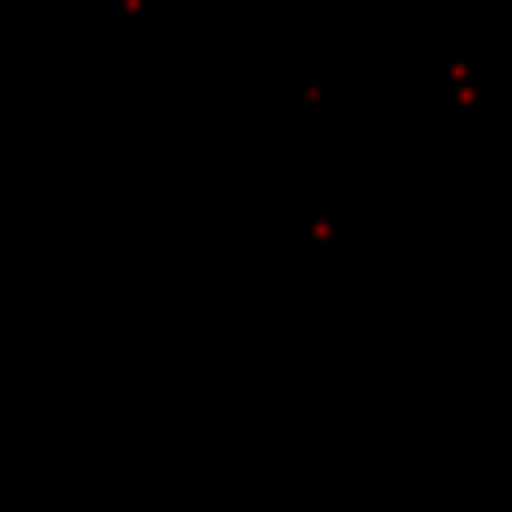} & 
            \includegraphics[width=\mywbsb]{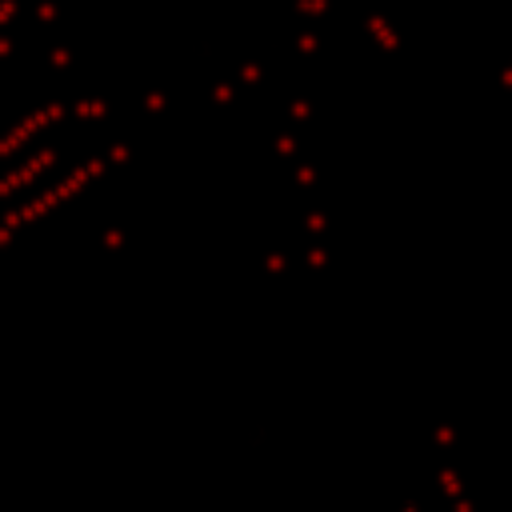} & 
            \includegraphics[width=\mywbsb]{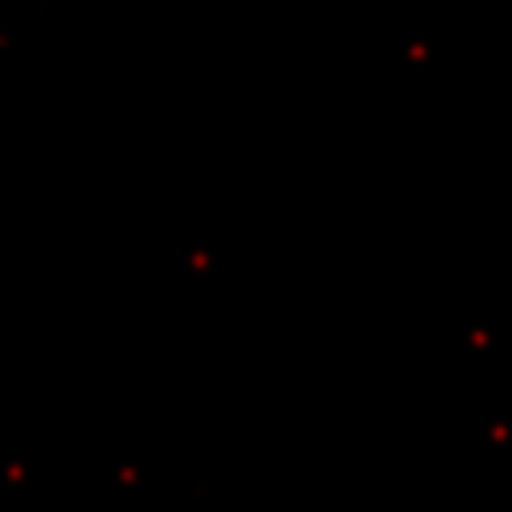} & 
            \includegraphics[width=\mywbsb]{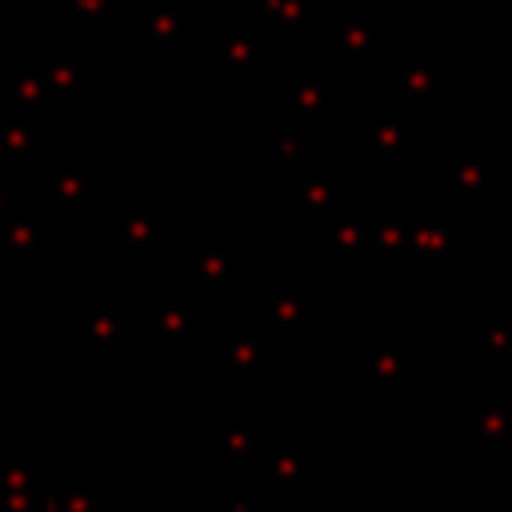}
            \\[-0.2em]

            \includegraphics[width=\mywbsb]{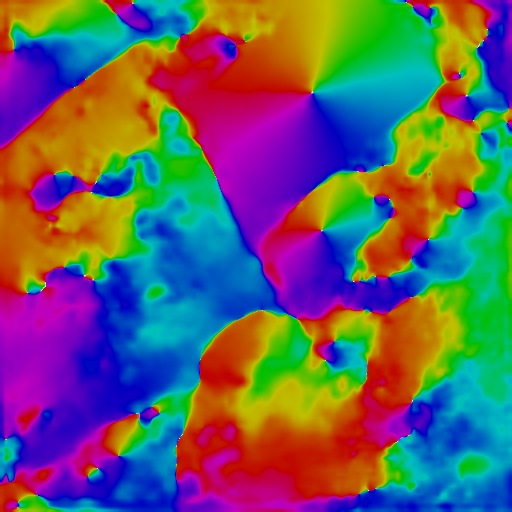} & 
            \includegraphics[width=\mywbsb]{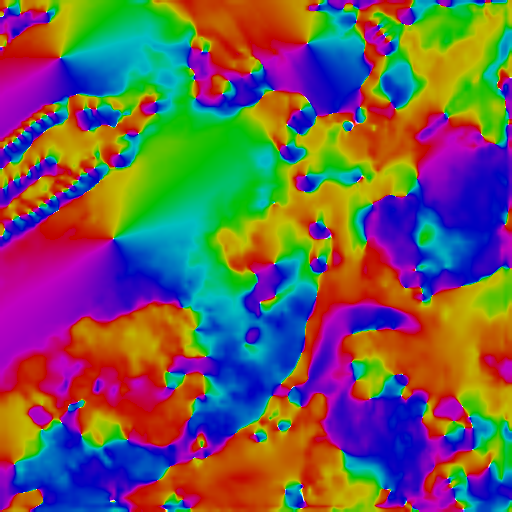} & 
            \includegraphics[width=\mywbsb]{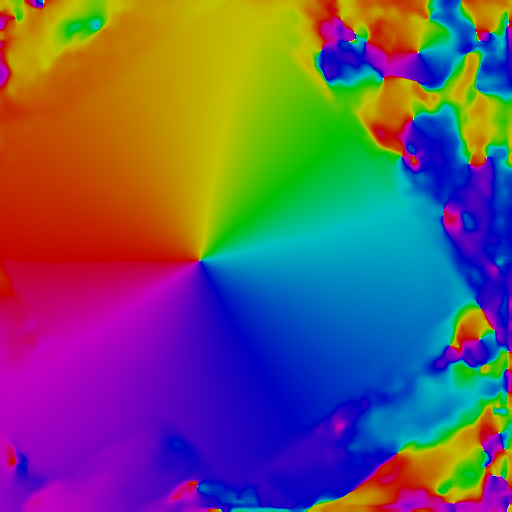} & 
            \includegraphics[width=\mywbsb]{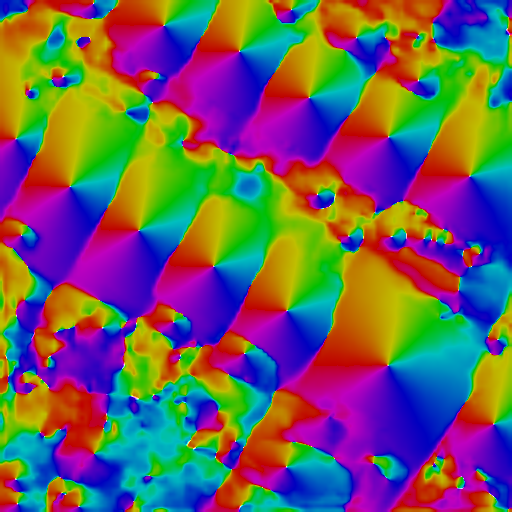}
            \\[-0.2em]

            \includegraphics[width=\mywbsb]{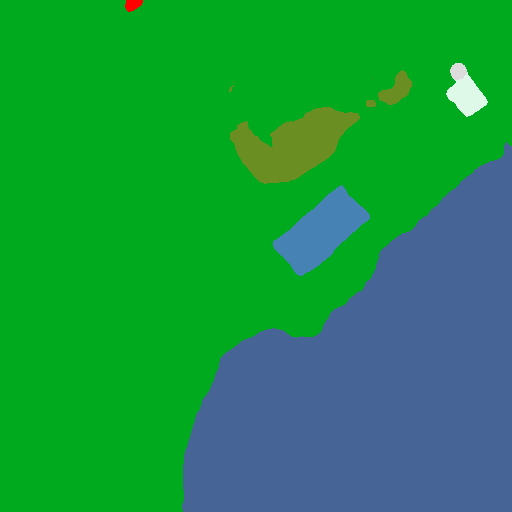} & 
            \includegraphics[width=\mywbsb]{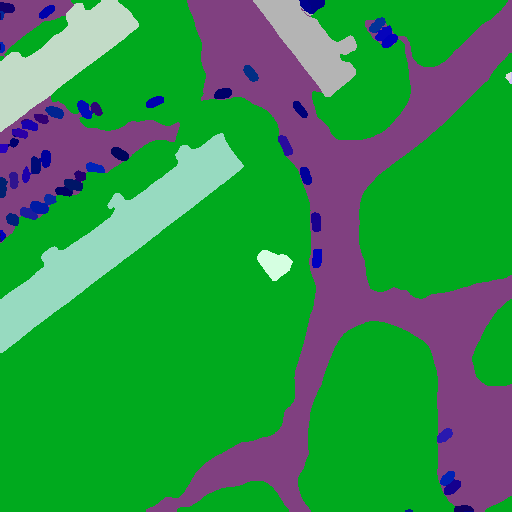} & 
            \includegraphics[width=\mywbsb]{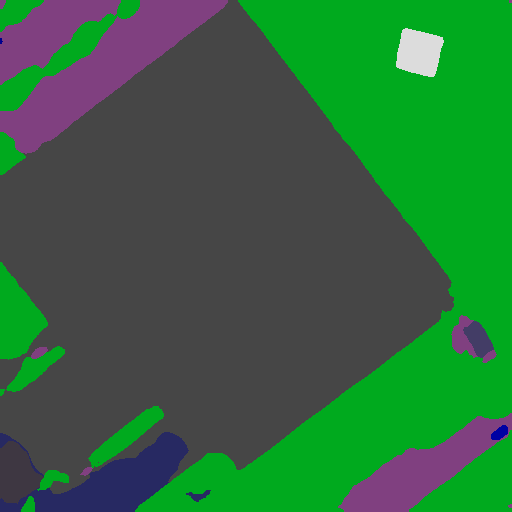} & 
            \includegraphics[width=\mywbsb]{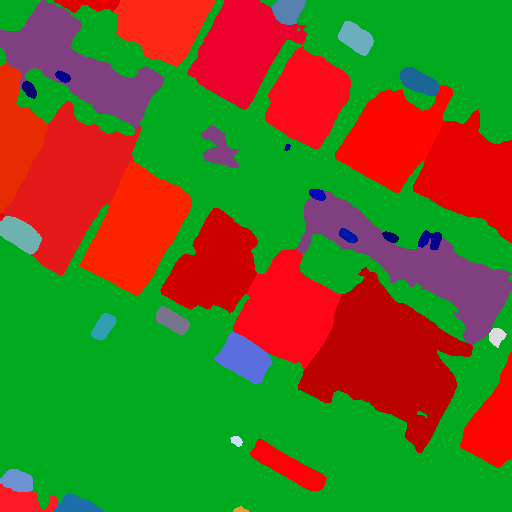}
            \\[-0.2em]
    \end{tabular}
    \caption{Qualitative results on four scenes from BSB val. Rows show: i) input image, ii) semantic segmentation, iii) centers heatmap, iv) offset directions, and v) panoptic segmentation.}
    \label{fig:bsb_preds}
\end{figure}

\vspace{0.15cm}
\subsection{Cityscapes}
The Cityscapes dataset contains 2975 training, 
500 validation and 1525 test images.
All images are densely labeled.
We train the models for 90000 iterations
with the ADAM optimizer
on $1024 \times 2048$ crops with batch size 8.
This can be carried out on 2 GPUs,
each with 20GiB of RAM.
We augment the crops through
horizontal flipping
and random scaling with factor between 0.5 and 2.0.
As before, we decay the learning rate from
$5 \cdot 10^{-5}$ to $10^{-7}$.

Table~\ref{tab:main_res} compares
our panoptic performance and inference speed
with other methods from the literature.
Panoptic SwiftNet based on ResNet-18 (PSN-RN18)
delivers competitive panoptic performance
with respect to models with more capacity,
while being significantly faster.
In comparison with
Panoptic Deeplab \cite{Cheng_2020_CVPR}
based on MobileNet v2,
we observe that our method
is more accurate in all three subtasks
and also faster for $61\%$ when 
measured in the same runtime environment.
We observe that methods based on Mask R-CNN
\cite{Porzi_2019_CVPR,kirillov2019panopticfpn}
achieve poor inference speed and non-competitive
semantic segmentation performance 
in spite of a larger backbone.

\begin{table}[h]
    \centering
        \caption{
    Panoptic segmentation performance on Cityscapes val.
    Sections correspond to top-down methods 
    based on Mask R-CNN (top), 
    bottom up methods (middle), 
    and our methods based on pyramidal fusion (bottom).
    FPS denotes frames per second from PyTorch and Python.
    FPS* denotes frames per second with FP16 TensorRT optimization. 
    $\dagger$ --- denotes our measurements.}
    \begin{tabular}{lcccccc}
         Model & PQ & AP & mIoU & FPS & FPS* & GPU \\
         \toprule
         PanopticFPN-RN101 \cite{kirillov2019panopticfpn} & 58.1 & 33.0 & 75.7 & - & - & - \\ 
         UPSNet \cite{xiong2019upsnet} & 59.3 & 33.3 & 75.2 & 4.2 & - & GTX 1080Ti\\
         SeamLess-RN50 \cite{Porzi_2019_CVPR} & 60.9 & 33.3 & 74.9 & 6.7 & - &  V100 \\
         RTPS-RN50 \cite{Hou_2020_CVPR} & 58.8 & 29.0 & 77.0 & 10.1 & 30.3 & V100 \\ 
         \midrule
         DeeperLab-WMNv2 \cite{yang2019deeperlab} & 52.2 & - & - &  4.0 & - & V100 \\
         PDL-MNv3 \cite{Cheng_2020_CVPR} & 55.4  & - & - & 15.9 & - & V100  \\
         PDL-MNv2 \cite{Cheng_2020_CVPR} & 55.1  & 23.3 & 75.8 & 17.8$\dagger$ & 42.2$\dagger$ & RTX3090  \\
         \midrule
         PSN-RN18 (ours) & 55.9 & 26.1  & 77.2 & \textbf{28.6} & \textbf{61.7} & RTX3090 \\
    \end{tabular}
    \label{tab:main_res}
\end{table}

\subsection{COCO}
COCO dataset \cite{lin2014microsoft} contains
digital photographs collected by Flickr users.
Images are annotated 
with 133 semantic categories \cite{caesar2018coco}.
The standard split proposes 118K images
for training, 5K for validation
and 20K for testing.
We train for 200K iterations
with ADAM optimizer
on $640\times640$ crops
with batch size 48.
This can be carried out on 2 GPUs,
each with 20 GiB of RAM.
We set the learning rate to 
$4 \cdot 10^{-4}$
and use polynomial rate decay.

Table \ref{tab:coco} presents
our results on the COCO validation set.
We compare our performance
with previous work
by evaluating $640\times640$ crops.
We achieve the best accuracy
among the methods 
with real-time execution speed.
Although designed for 
large-resolution images,
pyramidal fusion performs competitively even
on COCO images with median resolution 
of $640 \times 480$ pixels.

\begin{table}[h]
    \centering
    \setlength{\tabcolsep}{3pt}
    \caption{Panoptic segmentation performance on COCO val. We do not optimize our models with TensorRT
    in order to ensure a fair comparison 
    with the previous work.
    Sections correspond to top-down methods 
    based on Mask R-CNN (top), 
    bottom-up methods (middle), 
    and our method based on pyramidal fusion (bottom).
    }
    \begin{tabular}{lccccc}
         Model & PQ & PQ\textsubscript{th} & PQ\textsubscript{st} & FPS & GPU \\
         \toprule
         PanopticFPN-RN101 \cite{kirillov2019panopticfpn} & 40.3 & 47.5 & 29.5 & - & - \\
         RTPS-RN50 \cite{Hou_2020_CVPR} & 37.1 & 41.0 & 31.3 & 15.9 & V100 \\
         \midrule
         DeeperLab-WMNv2 \cite{yang2019deeperlab} & 28.1 & 30.8 & 24.1 & 12.0 & V100 \\
         Max-Deeplab-S \cite{Wang_2021_CVPR} & 48.4 & 53.0 & 41.5 & 14.9 & V100 \\
         PanopticFCN-RN50 \cite{Li_2021_CVPR} & 42.8 & 47.9 & 35.1 & 16.8 & V100 \\
         PDL-MNV3 \cite{Cheng_2020_CVPR} & 30.0 & - & - & 26.3 & V100 \\
         \midrule
         PSN-RN18 (ours)  & 30.5 & 32.1 & 28.1 & 64.5 & RTX3090 \\ 
    \end{tabular}
    \label{tab:coco}
\end{table}

\subsection{Mapillary Vistas}
Mapillary Vistas dataset \cite{neuhold2017mapillary}
collects high-resolution images
of road driving scenes
taken under a wide variety of conditions.
It contains 18K train, 2K validation
and 5K test images
densely labeled
with 65 semantic categories.
We train for 200K iterations 
with ADAM optimizer.
We increase the training speed 
with respect to the Cityscapes configuration
by reducing the crop size to $1024 \times 1024$
and increasing the batch size to 16.
We compose the training batches
by over-sampling crops of rare classes.
During evaluation
we resize the input image
so that the longer side
is equal to 2048 pixels
while maintaining 
the original aspect ratio 
\cite{Cheng_2020_CVPR}.
Similarly, during training
we randomly resize 
the input image
so that the mean resolution
is the same as in
in the evaluation.

Table~\ref{tab:mvd} presents
the performance evaluation on Vistas val.
Our model achieves 
comparable accuracy
w.r.t.\ literature,
while being much faster.
These models are slower than 
their Cityscapes counterparts 
from Table \ref{tab:main_res} 
for two reasons.
First, the average resolution of Vistas 
images in our training 
and evaluation experiments 
is $1536 \times 2048$ while 
the Cityscapes resolution 
is $1024 \times 2048$. 
Second, these models also have 
slower classification and 
post-processing steps due to 
larger numbers of classes 
and instances.

\begin{table}[h!]
    \centering
    \caption{Panoptic segmentation performance of pyramidal fusion model on Mapillary Vistas val.
    We report average FPS over all validation images. 
    }
    \begin{tabular}{lcccc}
         Model & PQ & mIoU & FPS & GPU \\
         \toprule
         DeeperLab-WMNV2 \cite{yang2019deeperlab} & 22.4 & / & 3.3 & V100 \\
         PDL-MNv3 \cite{Cheng_2020_CVPR} & 28.8  & / & 6.8 & V100  \\
         PDL-RN50 \cite{Cheng_2020_CVPR} & 33.3 & / & 3.5 & V100 \\
         \midrule
         PSN-RN18 (ours) & 28.6 & 47.8 & 17.0 &  RTX3090 \\
    \end{tabular}
    \label{tab:mvd}
\end{table}

\newcommand{\myh}{2.7cm}
\begin{figure}[h!]
    \centering
    \begin{tabular}{@{}c@{\,}c@{\,}c@{}}
            \includegraphics[height=\myh]{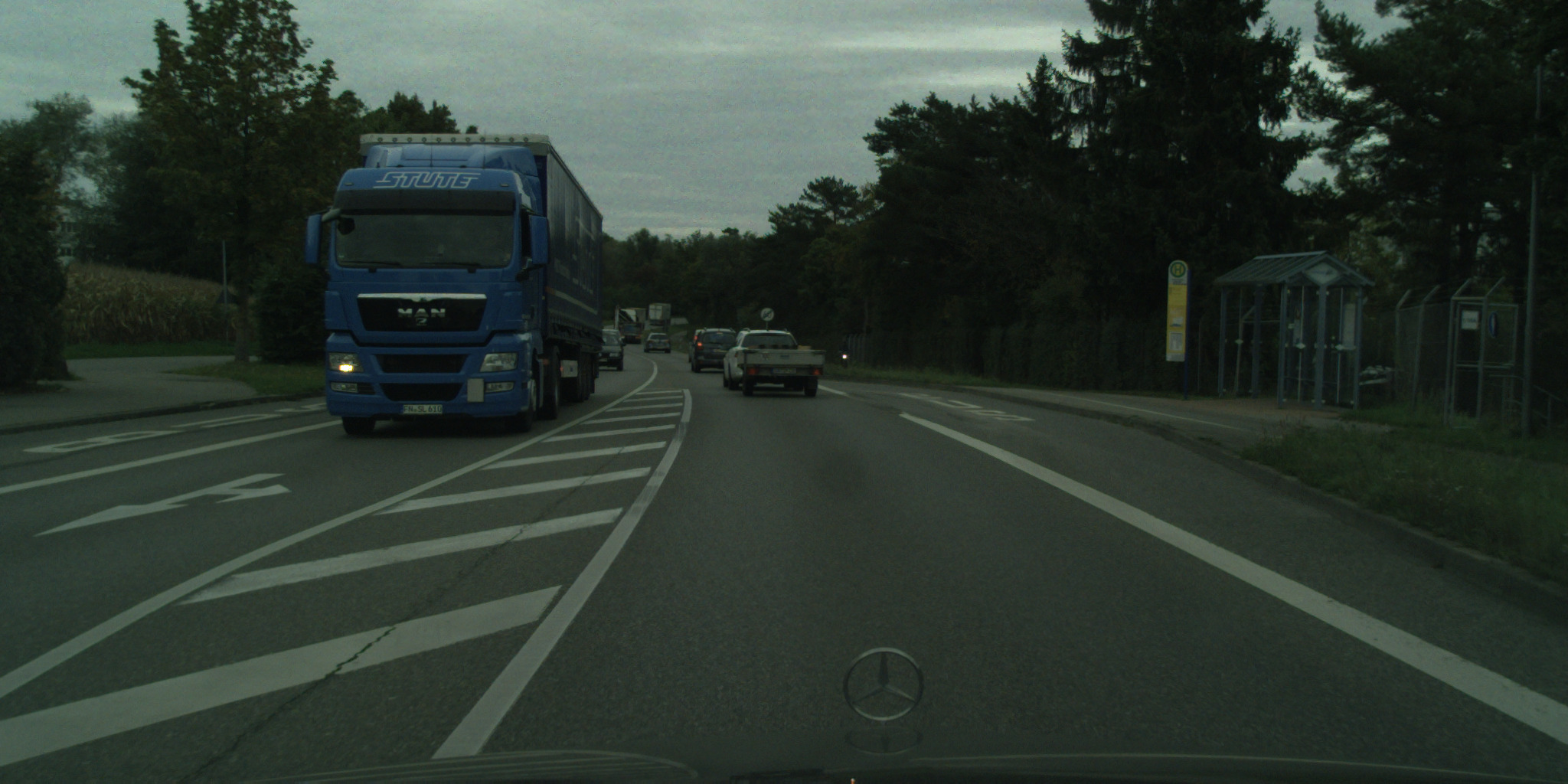} & 
            \includegraphics[height=\myh]{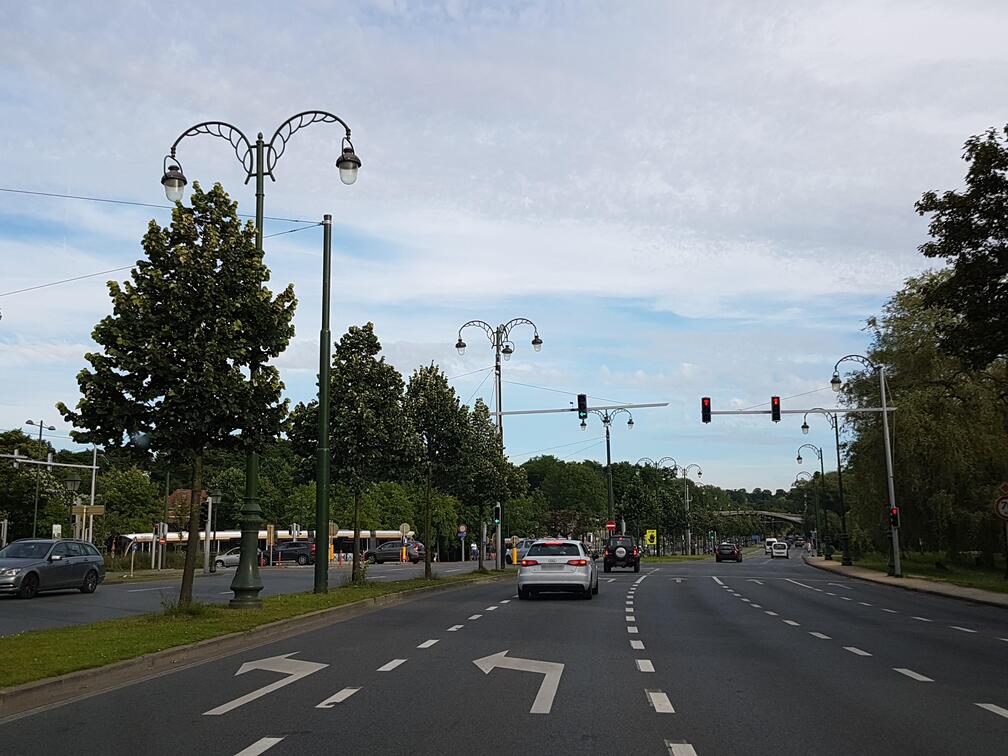} &
            \includegraphics[height=\myh]{coco-imgs/img_29.jpg} \\ [-0.2em]

            \includegraphics[height=\myh]{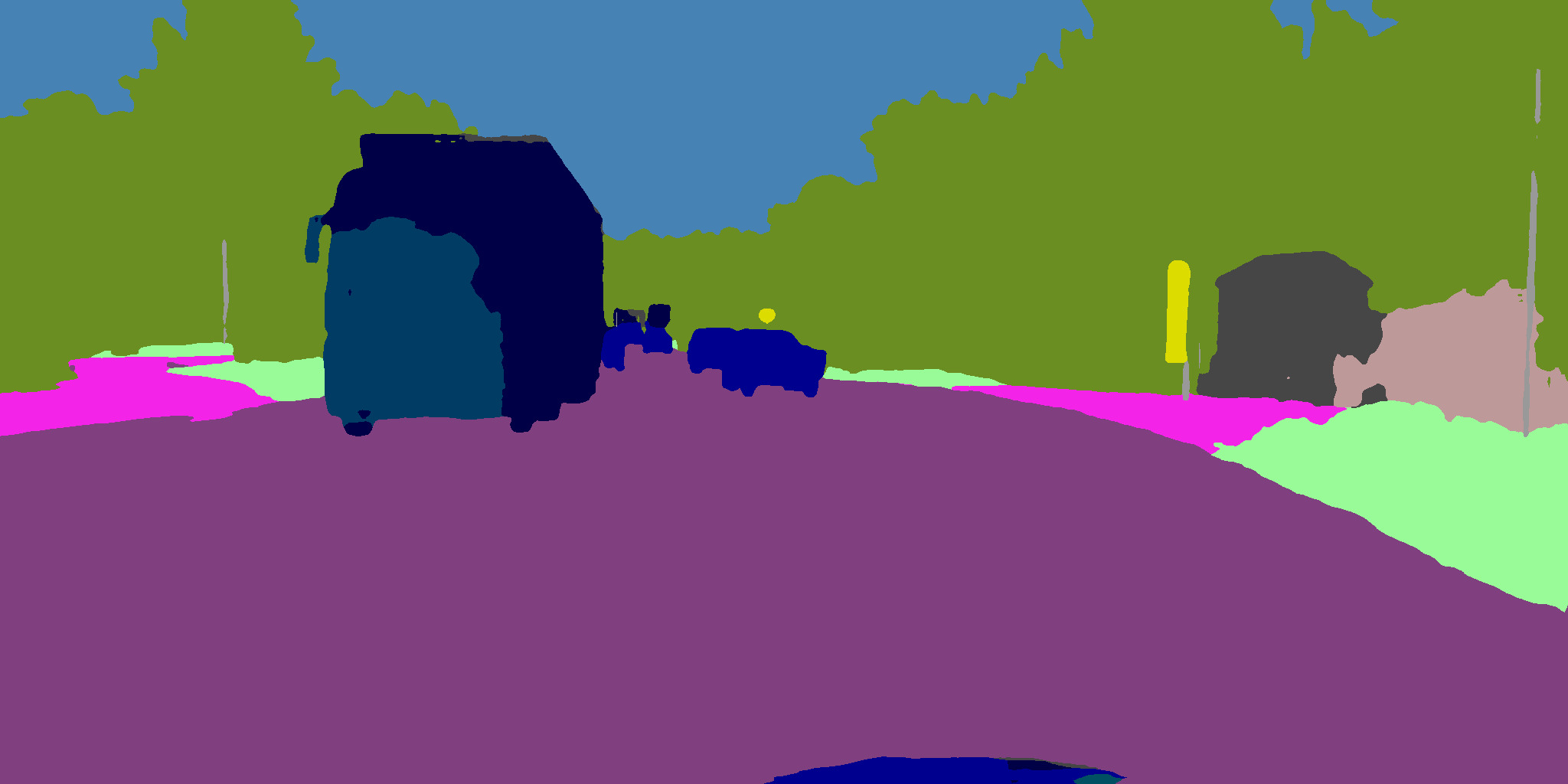} & 
            \includegraphics[height=\myh]{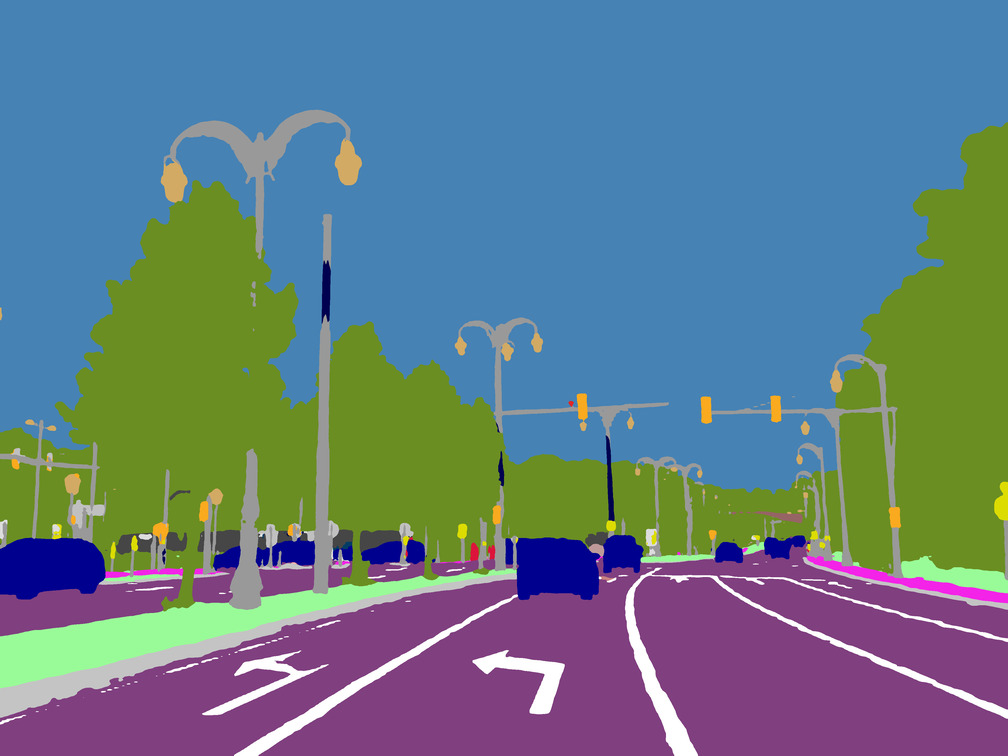} &
            \includegraphics[height=\myh]{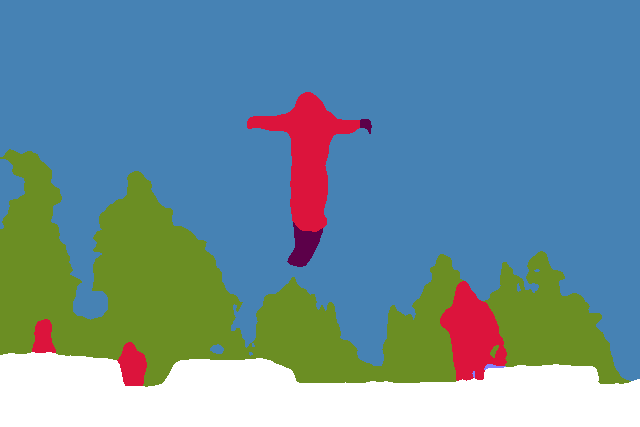} \\ [-0.2em]

            \includegraphics[height=\myh]{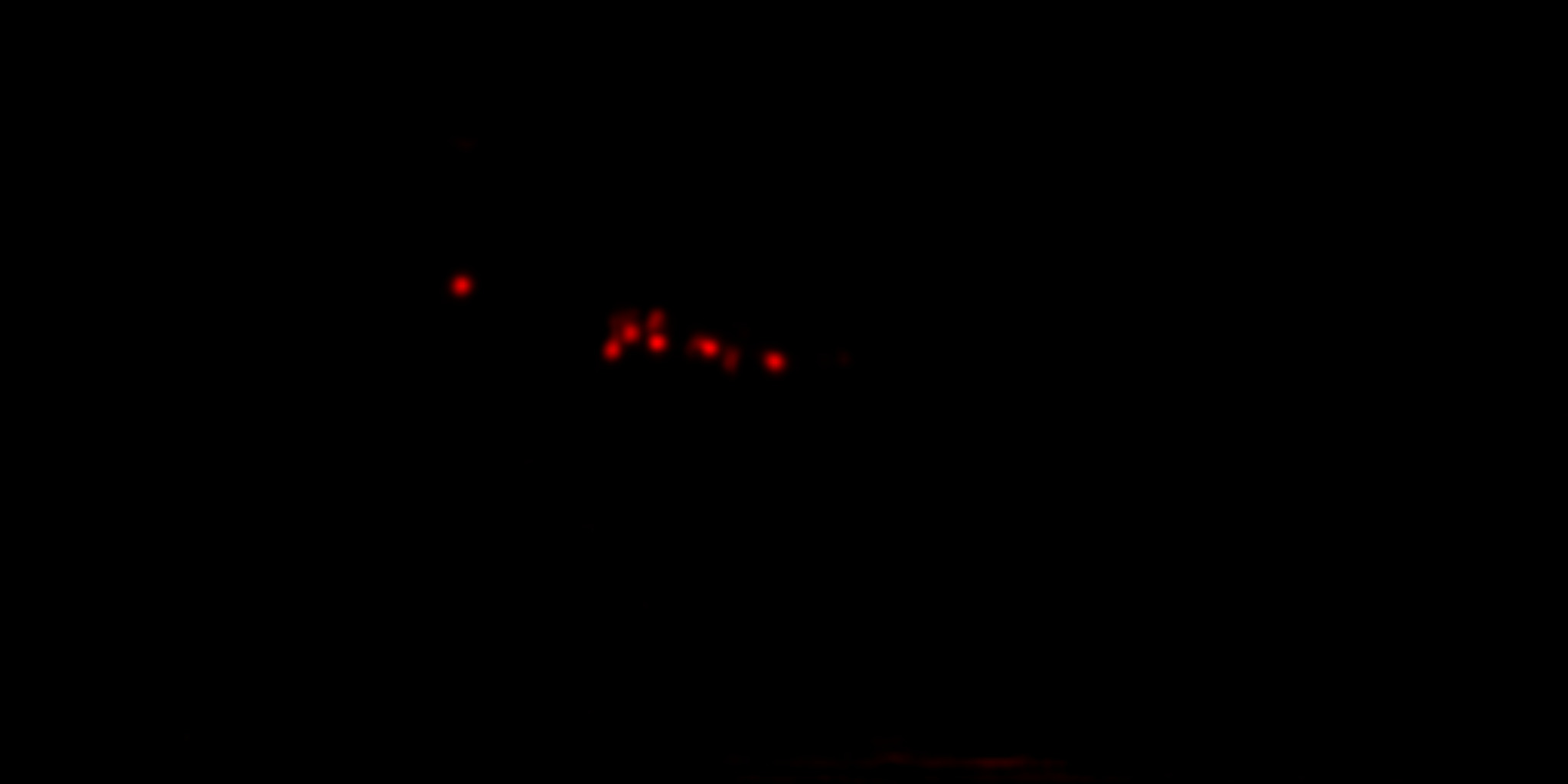} & 
            \includegraphics[height=\myh]{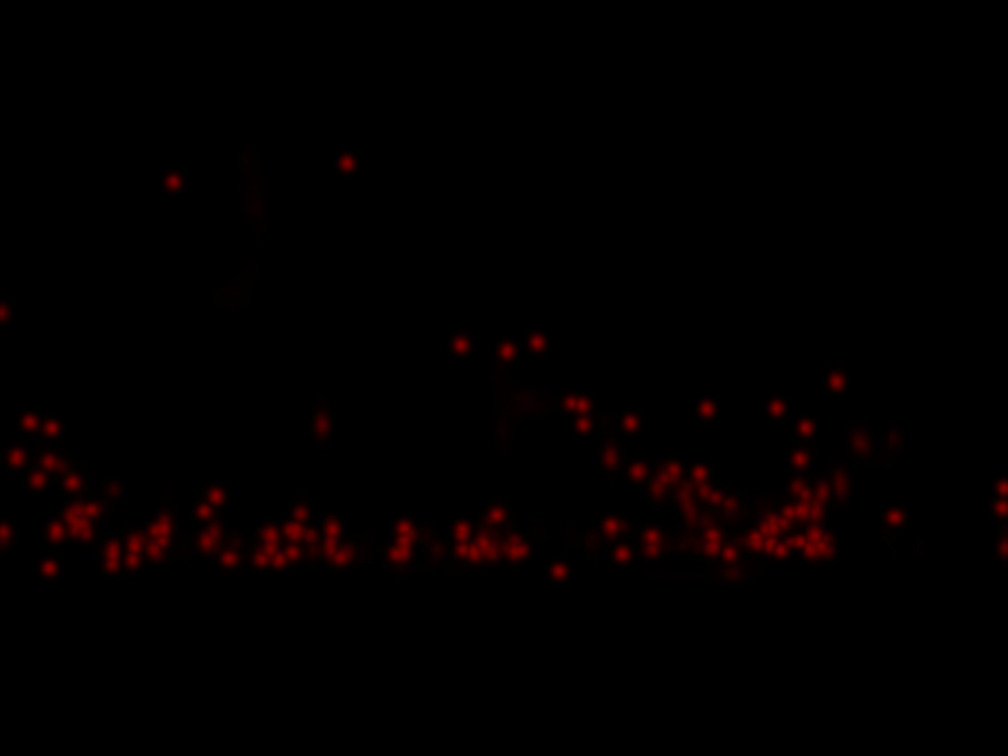} &
            \includegraphics[height=\myh]{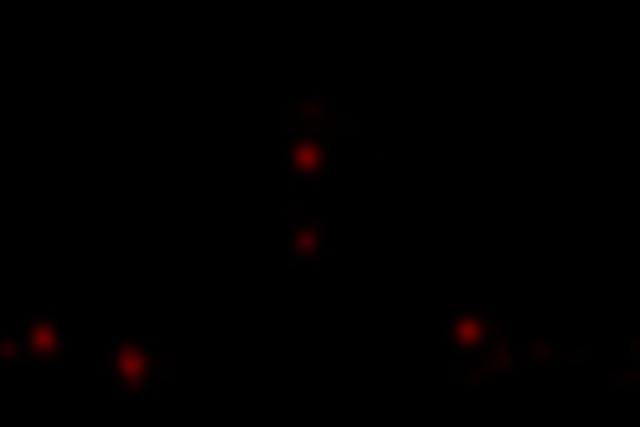} \\ [-0.2em]

            \includegraphics[height=\myh]{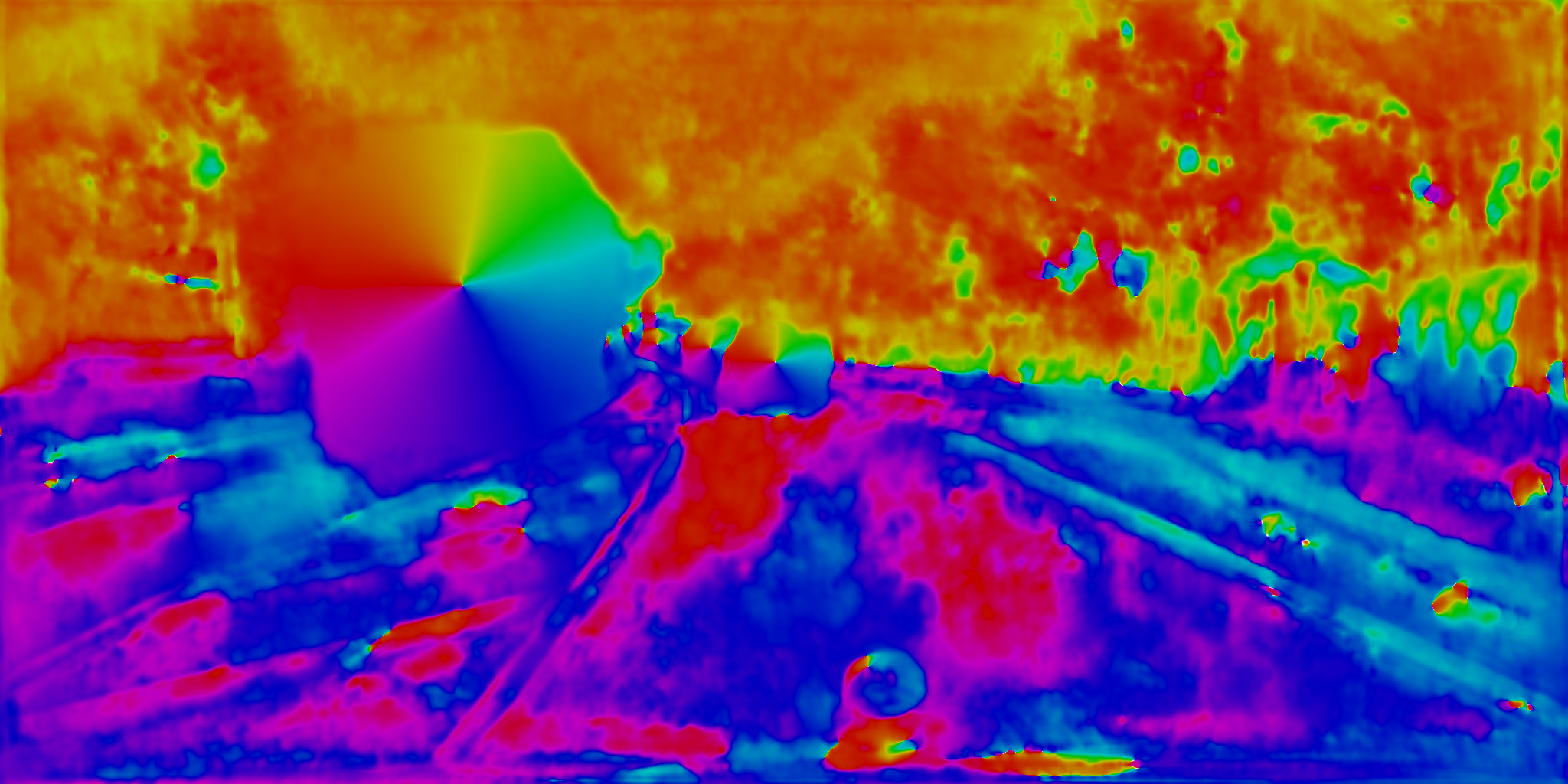} & 
            \includegraphics[height=\myh]{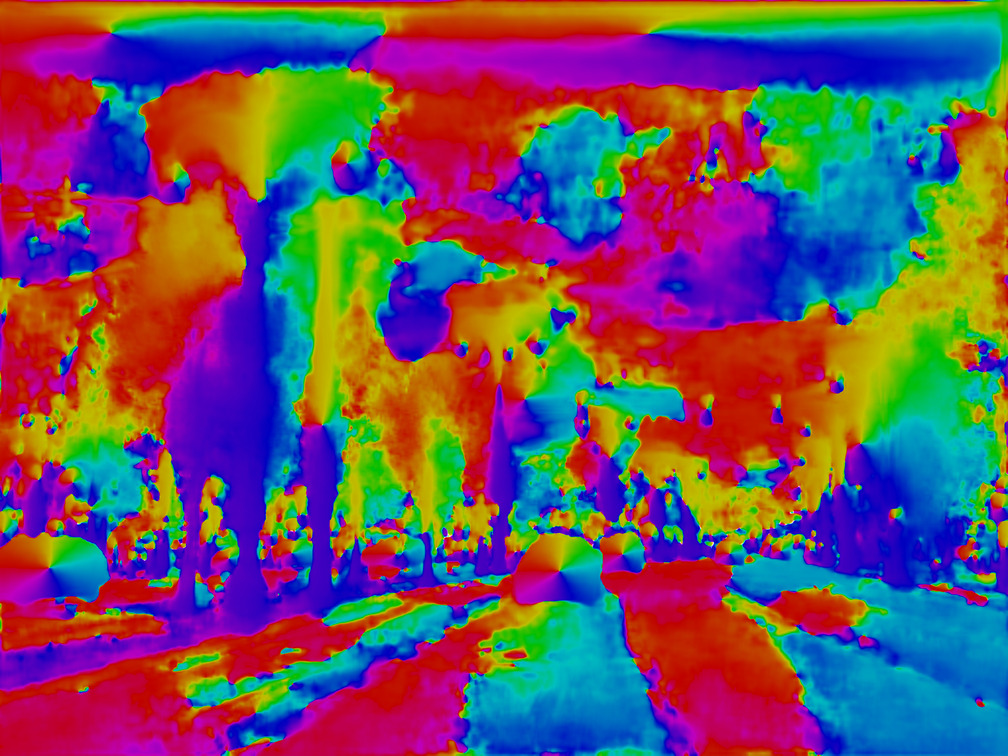} &
            \includegraphics[height=\myh]{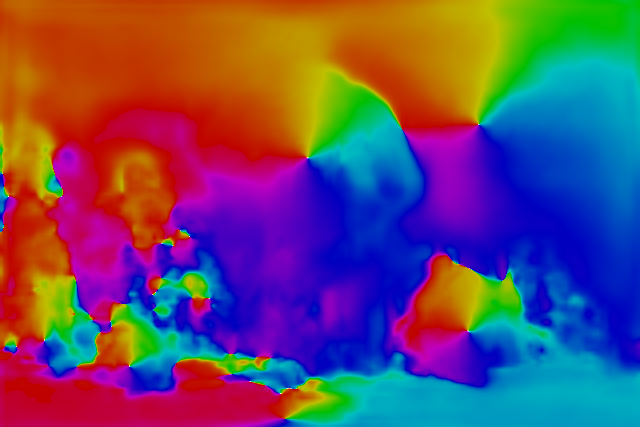} \\ [-0.2em]

            \includegraphics[height=\myh]{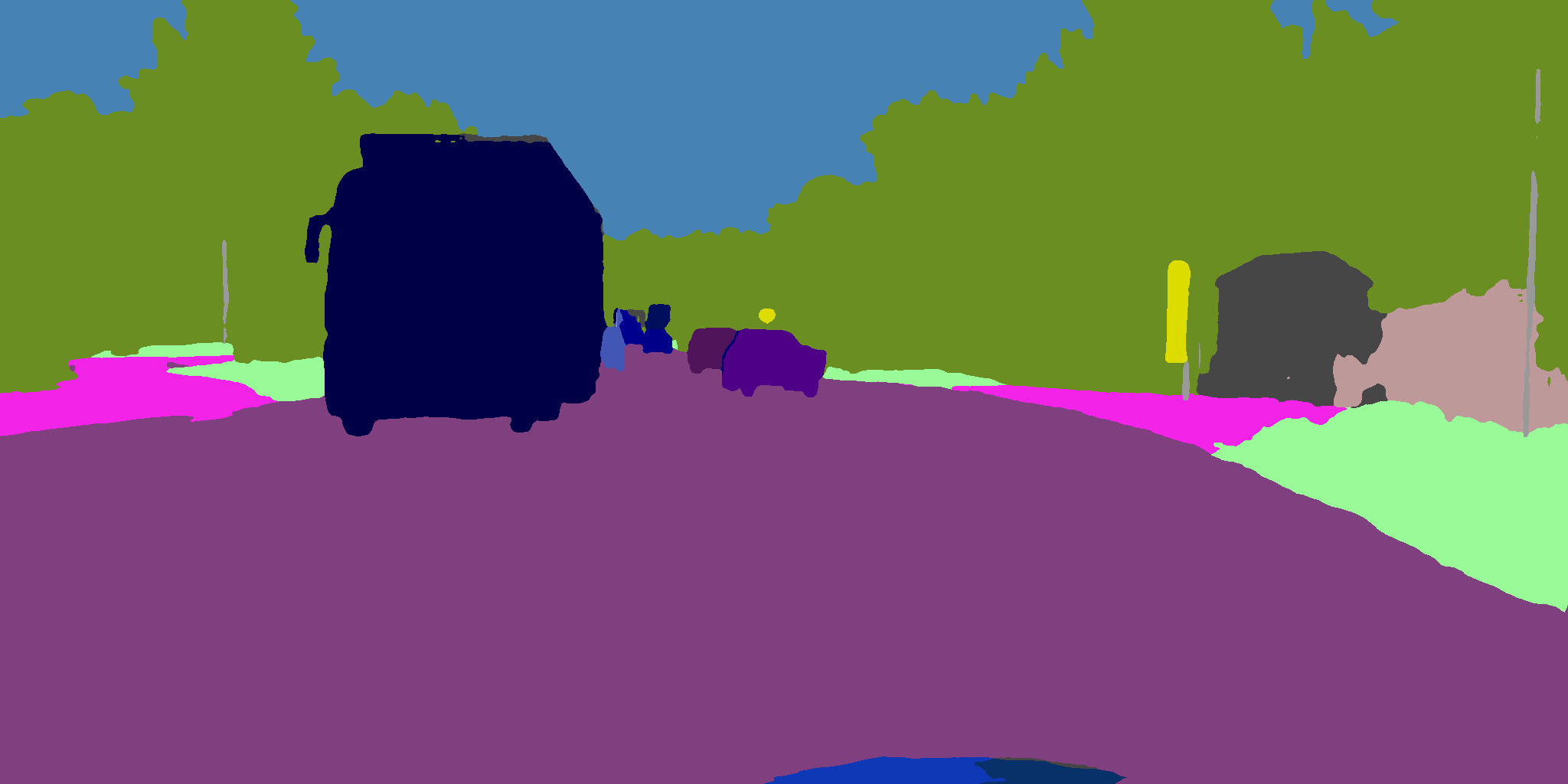} & 
            \includegraphics[height=\myh]{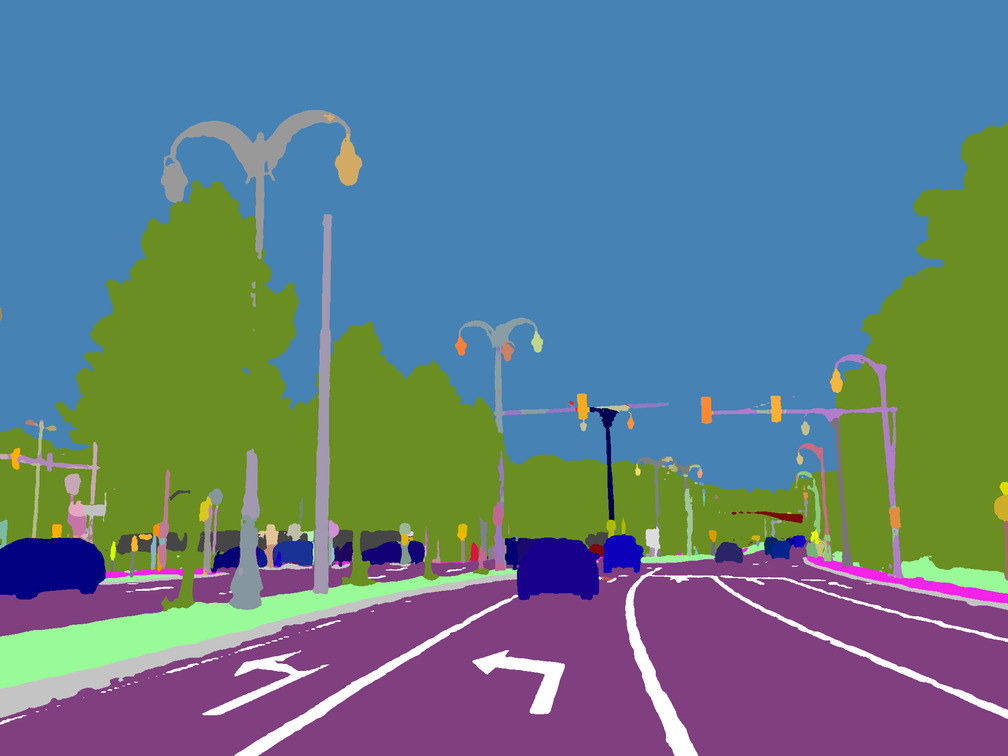} &
            \includegraphics[height=\myh]{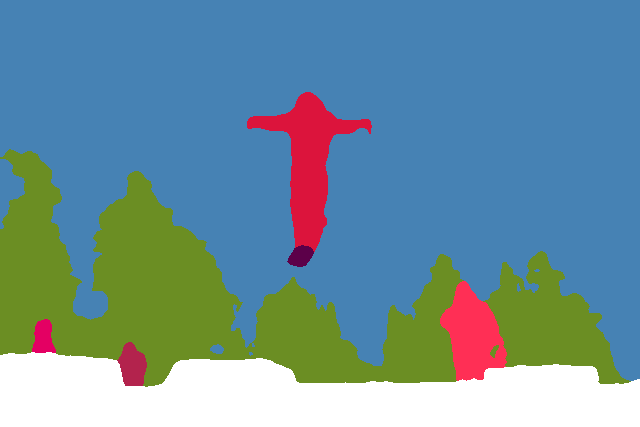} \\
    \end{tabular}
    \caption{Qualitative results on three scenes from Cityscapes, Mapillary Vistas and COCO. Rows show: i) input image, ii) semantic segmentation, iii) centers heatmap, iv) offset directions, and v) panoptic segmentation.}
    \label{fig:city-mvd-coco}
\end{figure}

\subsection{Qualitative experiments}
Figure~\ref{fig:city-mvd-coco} shows
qualitative results 
on three scenes from
validation sets of
Cityscapes, Mapillary Vistas and COCO.  
The rows show:
input image,
semantic segmentation,
centers heatmap,
offset directions
and the panoptic segmentation. 
Column 1 shows a Cityscapes scene
with a large truck on the left.
Semantic segmentation mistakenly recognizes
a blob of pixels in the bottom-left as class
bus instead of class truck.
However,
the panoptic map shows that the post-processing step 
succeeds to correct the mistake 
since the correct predictions outvoted the incorrect ones.
Hence, the instance segmentation of the
truck was completely correct.
Column 2 shows a scene
from Mapillary Vistas.
We observe that our model 
succeeds to correctly differentiate 
all distinct instances.
Column 3 shows a scene
from COCO val.
By looking at the offset predictions,
we can notice that the model
hallucinates an object
in the top-right part.
However,
this does not affect 
the panoptic segmentation
because that part of the scene
is classified as the sky 
(stuff class)
in semantic segmentation.
Thus, the offset predictions
in these pixels 
are not even considered.

\begin{figure}[h]
    \centering
    \newcommand{\pfpnwidth}{0.33\textwidth}
    \begin{tabular}{@{}c@{\,}c@{\,}c@{}}
         IMAGE & PFPN-R101 \cite{osmar2022panoptic,kirillov2019panopticfpn} & PSN-R18 (OURS) \\
         \includegraphics[width=\pfpnwidth]{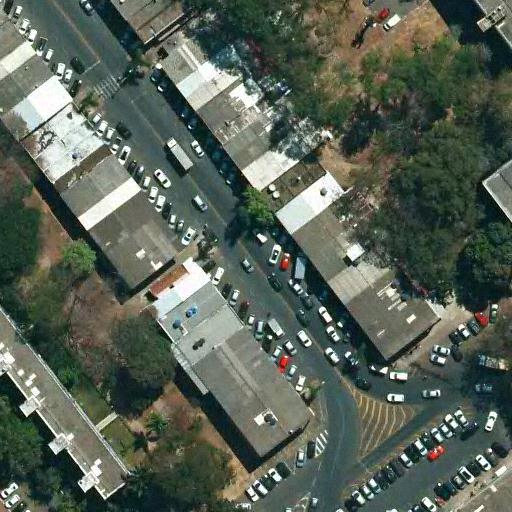} &
         \includegraphics[width=\pfpnwidth]{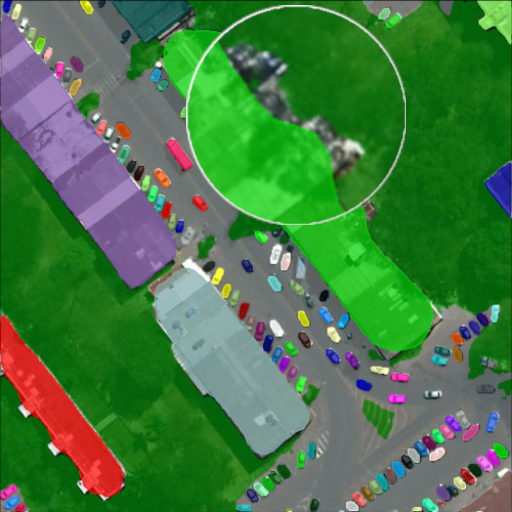} & 
         \includegraphics[width=\pfpnwidth]{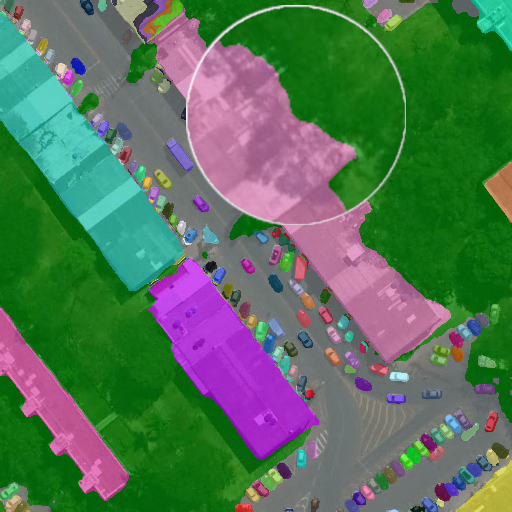} \\[-0.2em] 
         \includegraphics[width=\pfpnwidth]{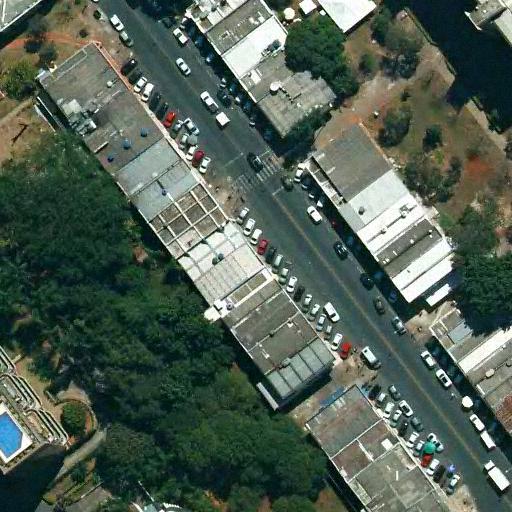} &
         \includegraphics[width=\pfpnwidth]{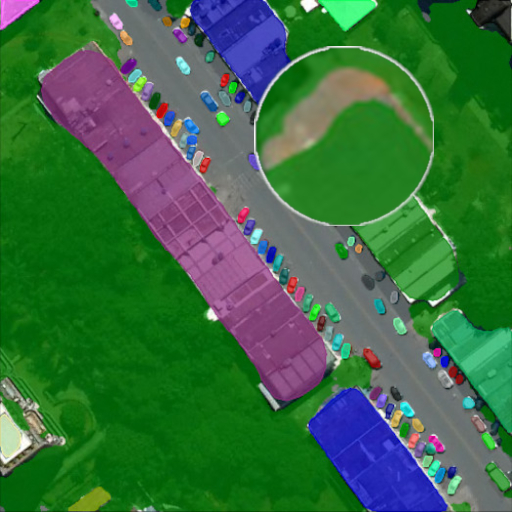} & 
         \includegraphics[width=\pfpnwidth]{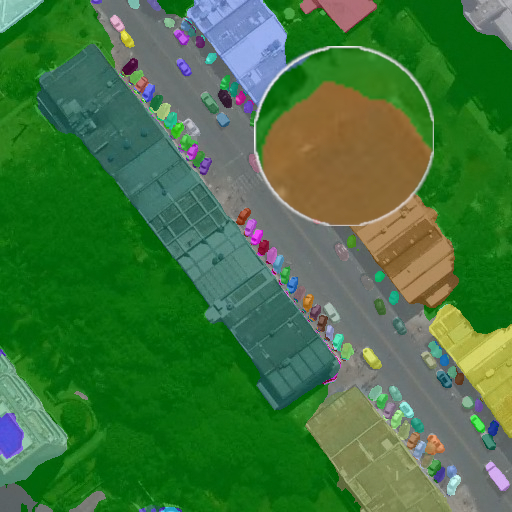} \\
         \includegraphics[width=\pfpnwidth]{pfpn_vs_psn/000000100238.jpg} &
         \includegraphics[width=\pfpnwidth]{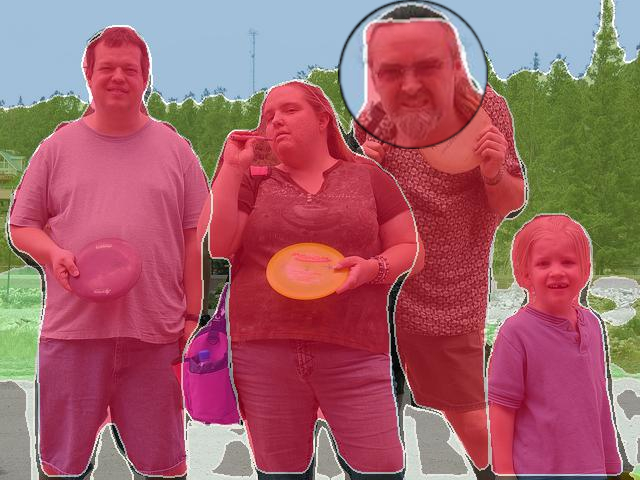} & 
         \includegraphics[width=\pfpnwidth]{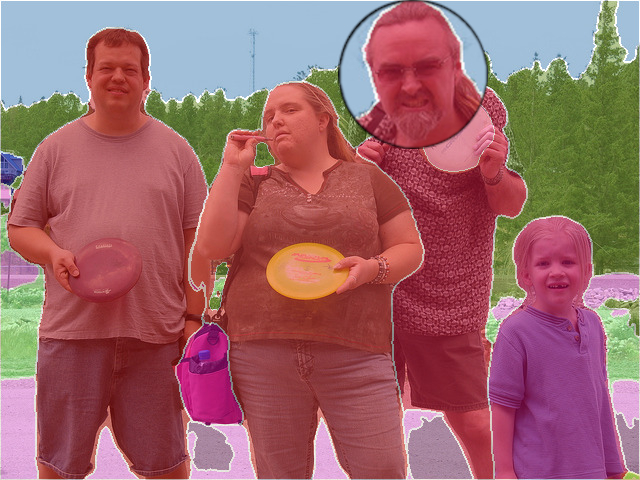} \\
         \includegraphics[width=\pfpnwidth]{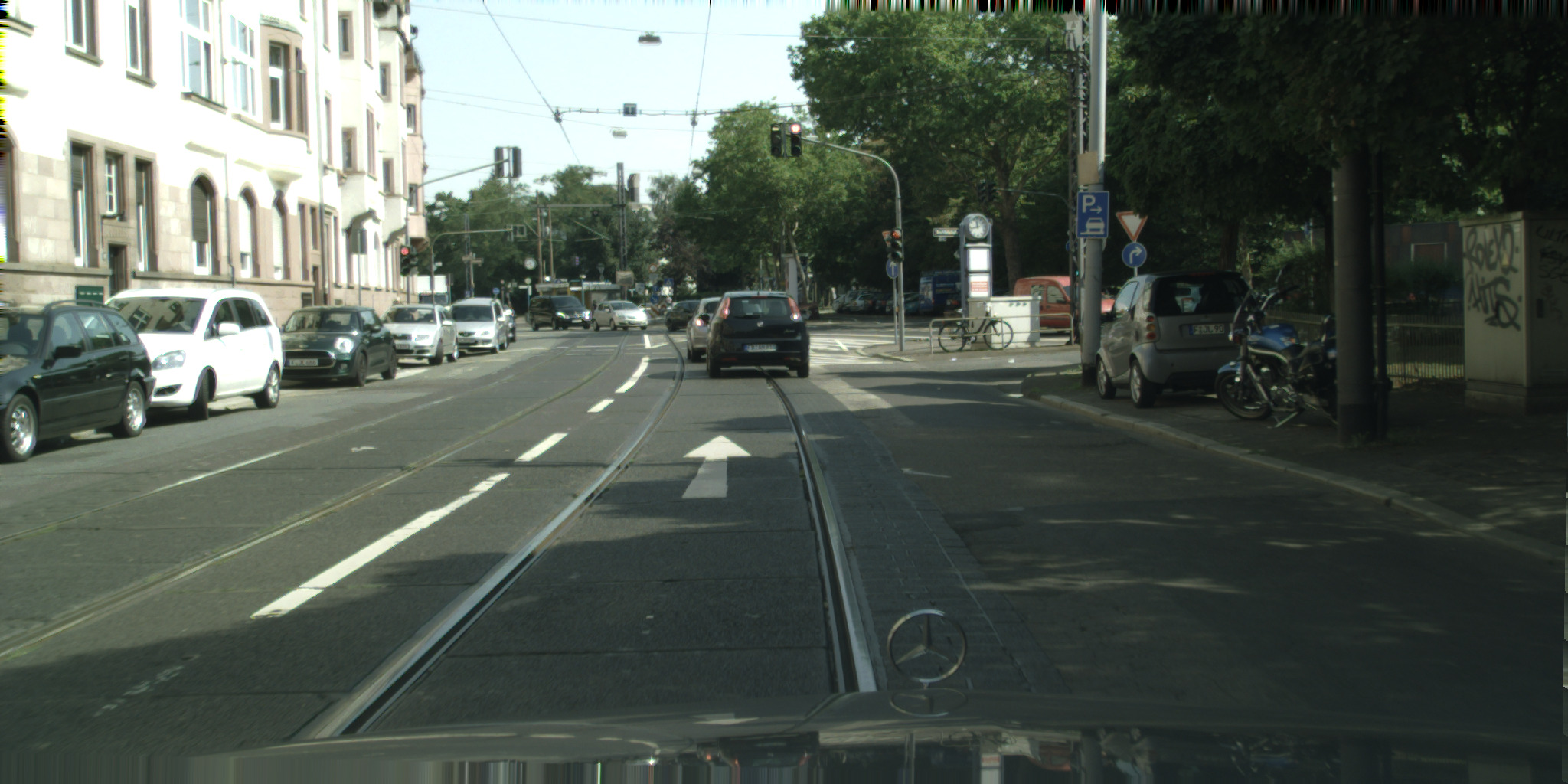} &
         \includegraphics[width=\pfpnwidth]{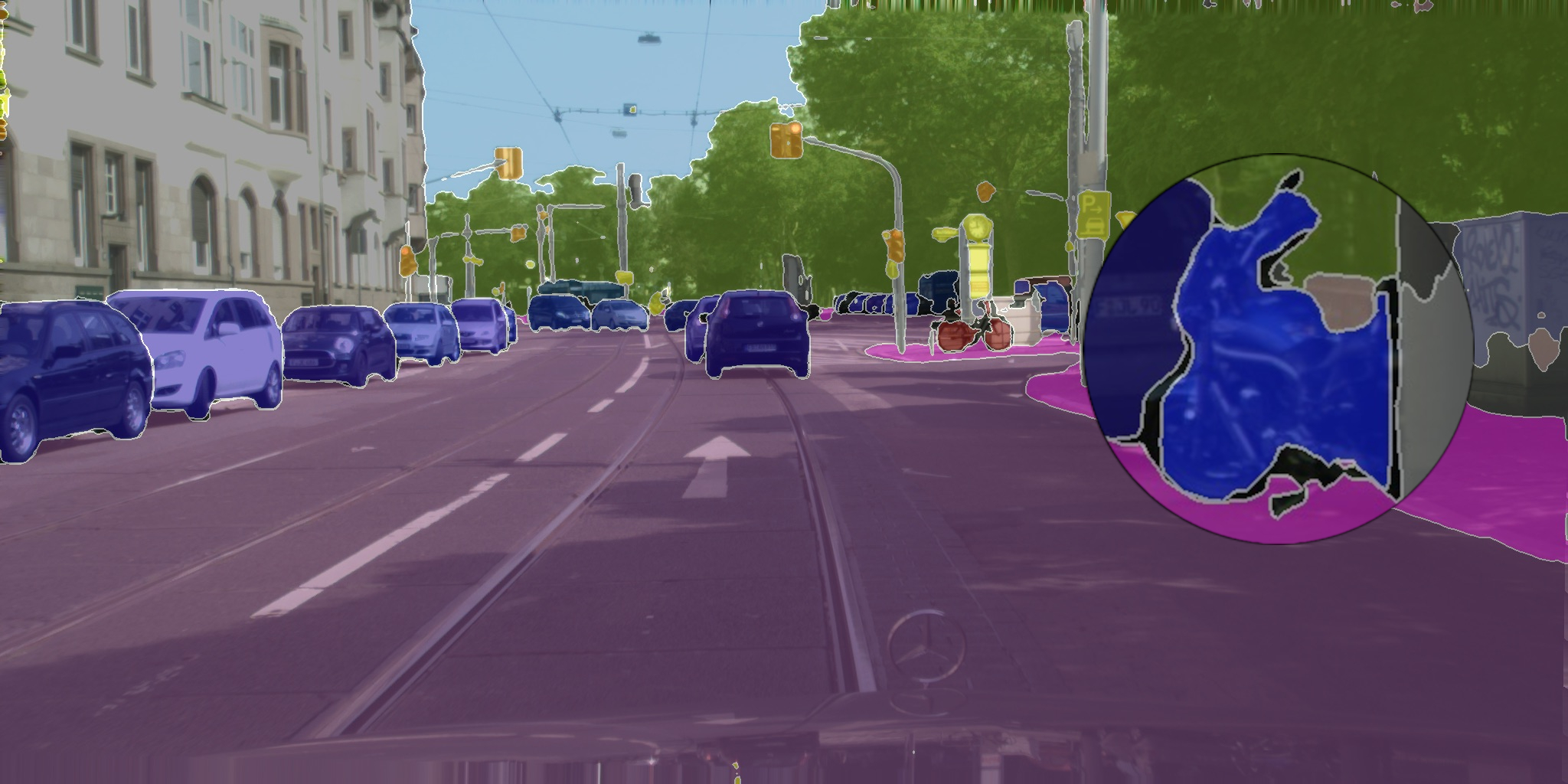} & 
         \includegraphics[width=\pfpnwidth]{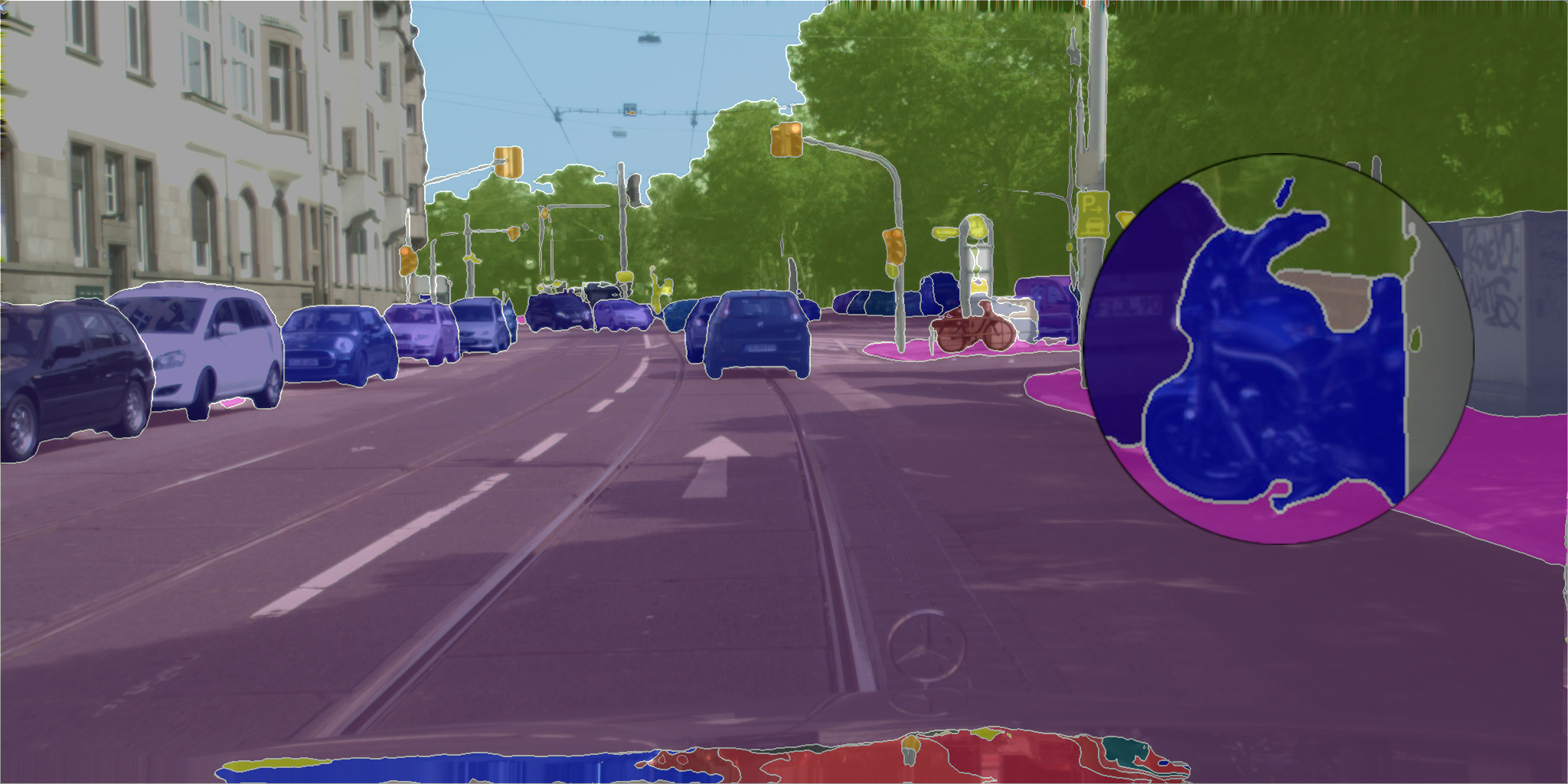} \\[-0.2em] 
    \end{tabular}
    \caption{
        Qualitative comparison between our method (column 3)
        and Panoptic FPN (column 2)
        on BSB Aerial dataset (rows 1 and 2), 
        COCO (row 3) and Cityscapes (row 4).
    }
    \label{fig:pfpn_comp}
\end{figure}

Figure~\ref{fig:pfpn_comp} presents 
a comparison  
between our predictions 
and those 
of the box-based 
Panoptic-FPN-R101.
The rows show
two scenes from 
the BSB Aerial dataset,
and one scene each
from COCO and Cityscapes.
The columns present the input image alongside 
the corresponding panoptic predictions.
We zoom in on 
onto circular image regions
where the differences between 
the two models are the most significant.
The Figure reveals
that our model
produces more accurate
instance masks
on larger objects.
Conversely, Panoptic FPN often misclassifies 
boundary pixels as background 
instead of a part of the instance.
This is likely due to 
generating instance segmentation 
masks on a small, 
fixed-size grid \cite{he2017mask}.
In contrast, our 
pixel-to-instance assignments are
trained on a fine resolution 
with a boundary-aware objective.
We also observe
that our model sometimes 
merges clusters of
small instances
in the distance.
We believe this is because
of the center-point-based 
object detection and
corresponding non-maximum suppression.
Box-based approaches
perform better 
in such scenarios,
which is consistent
with the AP evaluation
in Table~\ref{tab:main_res}.

\subsection{Validation study on the Cityscapes dataset}
This study
quantifies the contribution
of pyramidal fusion 
and boundary-aware offset loss
to panoptic segmentation accuracy.
Table~\ref{tab:pyr_val} compares
pyramidal fusion with 
spatial pyramid pooling \cite{zhao2017pyramid,krevso2020efficient}
as alternatives for providing
global context.
We train three separate models
with 1, 2, and 3 levels of pyramidal fusion,
as well as a single-scale model with 
spatial pyramidal pooling 
of 32x subsampled features 
\cite{krevso2020efficient}.
We observe that spatial pyramid pooling 
improves AP and mIoU 
for almost 4 percentage points (cf. rows 1 and 2).
This indicates that 
standard models
are unable to capture global context
due to undersized receptive field.
This is likely exacerbated by the fact
that we initialize with parameters obtained 
by training on $224 \times 224$ ImageNet images.
We note that two-level pyramidal fusion 
outperforms the SPP model 
across all metrics.
Three-level pyramid fusion achieves 
further improvements
across all metrics. 
PYRx3 outperforms PYRx2 
for 1 PQ point, 2.8 AP points, and 1.1 mIoU points.

\begin{table}[h]
    \centering
        \caption{
        Comparison of 
        pyramidal fusion 
        with spatial pyramid pooling \cite{zhao2017pyramid}
        on Cityscapes val. 
        Both PYRx1 and SPP models
        operate on single resolution
        and hence can not exploit pyramidal fusion. 
        The SPP model applies spatial pyramid pooling
        to $32\times$ subsampled abstract features 
        at the far end of the backbone \cite{krevso2020efficient}.
    }
    \begin{tabular}{lccccccc}
         Method & PQ & SQ & RQ & AP & mIoU & FPS \\
         \toprule
        PSN-RN18 single & 51.7 & 78.4 & 64.3 & 18.7   & 71.2 & 34.5 \\
        PSN-RN18 SPP & 53.5 & 78.8 & 66.3 & 22.4  & 75.1 & 32.2 \\
        PSN-RN18 PYRx2 & 54.9 & 79.2 & 67.8 & 23.3 & 76.1 & 30.3 \\
        PSN-RN18 PYRx3 & \textbf{55.9} & \textbf{79.5} & \textbf{68.9} & \textbf{26.1}  & \textbf{77.2} & 28.6 \\
    \end{tabular}
    \label{tab:pyr_val}
\end{table}

Table~\ref{tab:oracle_eval} explores
the upper performance bounds
w.r.t.\ particular model outputs.
These experiments indicate that 
semantic segmentation represents 
the most important challenge 
towards accurate panoptic segmentation.
Perfect semantic segmentation
improves panoptic quality
for roughly 20 points.
However, we believe that rapid progress 
in semantic segmentation accuracy 
is not very likely due 
to wide popularity of the problem.
When comparing oracle center and oracle offsets
we observe that the latter one
\begin{table}[h]
    \centering
        \caption{Evaluation of models with different oracle components. Labels +GT X denote models which use groundtruth information for the prediction X. Note that c\&o\ denotes centers and offsets.}
    \begin{tabular}{lcccccc}
         Model & PQ & SQ & RQ & AP & mIoU \\
         \toprule
         PSN-RN18 & {55.9} & {79.5} & {68.9} & {26.1}  & {77.2} \\
         \midrule
         + GT semseg & {76.2} & {91.0} & {82.5} & {39.6}  & 100.0 \\
         + GT centers & {56.9} & {79.5} & {70.1} & {22.9}  & 77.2 \\
         + GT offsets & {59.1} & {83.0} & {70.4} & {36.8}  & 77.2 \\
         + GT c\&o & 61.6 & 83.3 & 73.2 & 38.2 & 77.2 \\
    \end{tabular}
    \label{tab:oracle_eval}
\end{table}
brings 
significantly 
larger improvements:
3 PQ and 10 AP points over the regular model.
In early experiments 
most of the offset errors
were located at instance boundaries,
so we introduced the boundary-aware offset loss 
(\ref{eq:bal}).

Table~\ref{tab:l1_bal} explores the influence 
of the boundary-aware offset loss (\ref{eq:bal})
to the PSN-18 accuracy.
We consider three variants
based on the number of regions
that divide each instance.
In each variant,
we set the largest weight 
equal to 8
for the region closest 
to the border,
and then gradually reduce it
to 1 for the most distant region.
Note that we set the overall 
weight of the offset loss to 
$\lambda=0.01$
when we do not use the
the boundary-aware formulation (\ref{eq:bal}).
We observe that the 
boundary-aware loss
with four regions per instance
brings noticeable 
improvements across all metrics.
The largest improvement is 
in instance segmentation performance 
which increases for 1.1 AP points.

\begin{table}[h]
    \centering
        \caption{Validation of boundary-aware offset loss for the PSN-RN18 PYRx3 model on Cityscapes val. 
        The presented results are average over two runs.}
    \begin{tabular}{lcccccc}
         Offset loss & PQ & SQ & RQ & AP & mIoU \\
         \toprule
         L1 & 55.2 & 79.2 & 68.2 & 25.2  & 76.3 \\
         L1-BAL (3 regions per instance) & 55.2 & 79.2 & 68.2 & 25.5 & 76.6 \\
         L1-BAL (4 regions per instance) & \textbf{55.9} & \textbf{79.5} & \textbf{68.9} & \textbf{26.3}  & \textbf{76.6} \\
         L1-BAL (8 regions per instance) & 55.5 & 79.3 & 68.5 & 25.2 & 76.8 \\
    \end{tabular}
    \label{tab:l1_bal}
\end{table}

\subsection{Inference speed after TensorRT optimization}
\begin{figure}[h]
    \centering
    \includegraphics[width=0.8\textwidth]{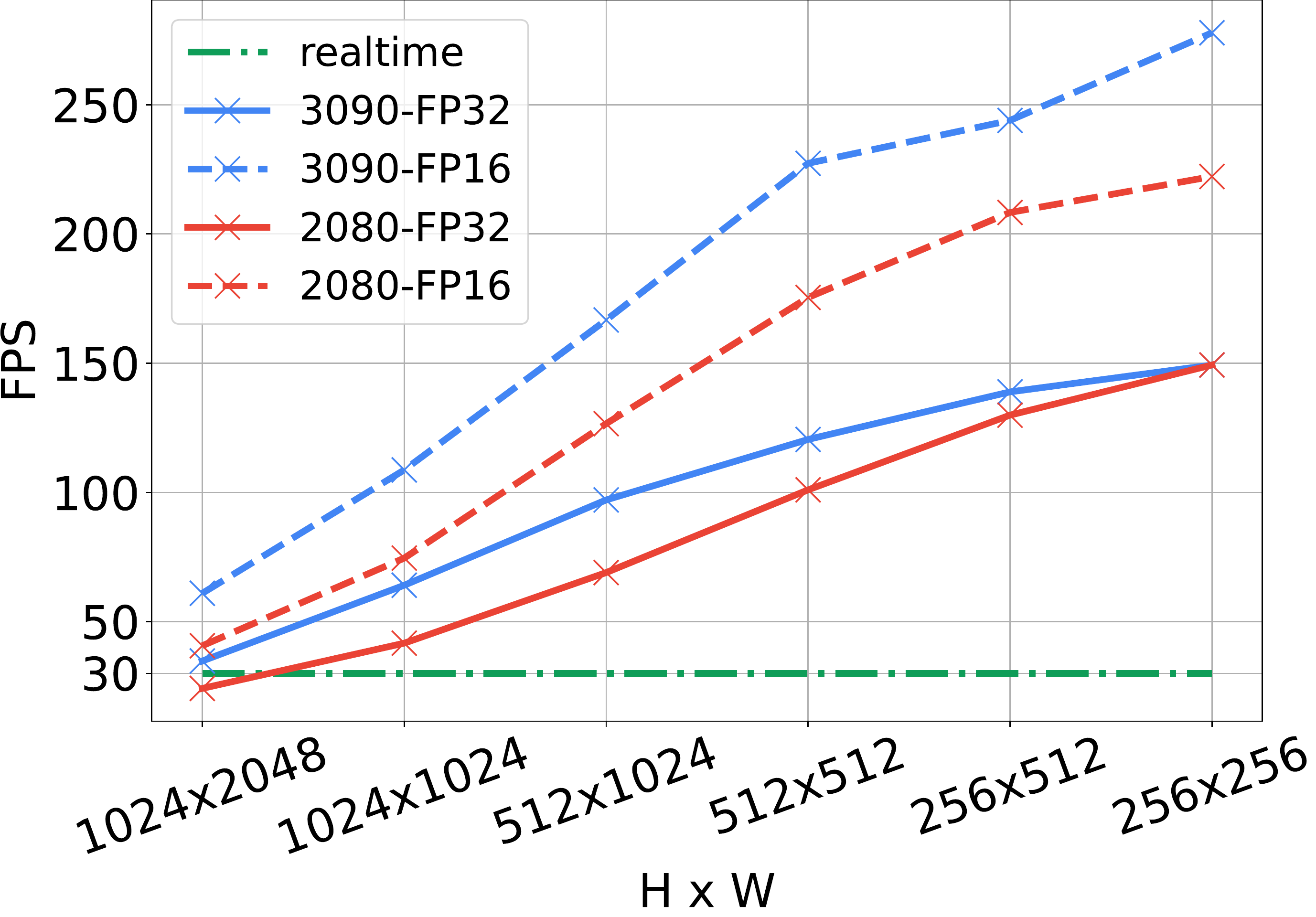}
    \caption{Inference speed of a three-level PSN-RN18 
    across different input resolutions
    for two graphics cards and two precisions.
    All configurations involve the same model
    after optimizing it with TensorRT. 
    All datapoints are averaged over
    all images from Cityscapes val
    in order to account for the dependence 
    of the post-processing time on scene complexity.
    }
    \label{fig:trt-runtime}
\end{figure}

Figure~\ref{fig:trt-runtime} evaluates 
speed of our optimized models 
across different input resolutions.
All models have been run from Python
within the TensorRT execution engine. 
We have used TensorRT to optimize the same model
for two graphics cards and two precisions (FP16 and FP32).
Note however that these optimizations
involve only the network inference,
while the post-processing step
is executed with a combination
of Pytorch \cite{paszke2019pytorch} 
and cupy \cite{cupy_learningsys2017}.

Interestingly, FP16 brings almost $2\times$ improvement
across all experiments, although RTX3090 
declares the same peak performance for FP16 and FP32. 
This suggests that 
it is likely that our performance bottleneck 
corresponds to memory bandwidth 
rather than computing power.

We start the clock when an image 
is synchronized in CUDA memory
and stop it when the post-processing is completed.
We capture realistic post-processing times
by expressing all datapoints 
as average inference speed
over all images from Cityscapes val.
The figure shows that our model
achieves real-time inference speed
even on high-resolution 2MPx images. 
Our PSN-RN18 model achieves more than 100 FPS 
on 1MPx resolution with FP16 RTX3090.

\section{Conclusion}
We have proposed 
a novel panoptic architecture 
based on
multi-scale features, 
cross-scale blending,
and bottom-up instance recognition.
Ablation experiments indicate clear advantages
of the multi-scale architectures
and boundary-aware learning
to the panoptic performance.
Experiments with oracle components suggest
that semantic segmentation represents
a critical ingredient of panoptic performance.
Panoptic SwiftNet-RN18 achieves 
state-of-the-art 
generalization performance 
and the fastest inference
on the BSB Aerial dataset.
The proposed method
is especially appropriate for 
remote sensing applications
as it is able to 
efficiently process 
very large resolution inputs.
It also achieves
state-of-the-art performance
on Cityscapes as well as 
competitive performance 
on Vistas and COCO 
among all models 
aiming at real-time inference.
Source code 
will be publicly available
upon acceptance.

\section*{Acknowledgments}
This work has been funded by Rimac Technology.
This work has also been supported by the Croatian Science Foundation, grant IP-2020-02-5851 ADEPT, and European Regional Development Fund, grant KK.01.1.1.01.0009 DATACROSS.
\bibliographystyle{abbrv}
\bibliography{ref}

\end{document}